\documentclass{article}

\usepackage{microtype}
\usepackage{graphicx}
\usepackage{booktabs} 
\usepackage{enumitem}
\usepackage{cprotect}
\usepackage{subcaption}
\usepackage{pifont}
\usepackage{tablefootnote}

\usepackage{hyperref}

\usepackage[preprint]{icml2026}

\usepackage{amsmath}
\usepackage{amssymb}
\usepackage{mathtools}
\usepackage{amsthm}

\usepackage[capitalize,noabbrev]{cleveref}

\theoremstyle{plain}

\theoremstyle{definition}

\theoremstyle{remark}

\usepackage[textsize=tiny]{todonotes}

\usepackage{listings}
\usepackage{tcolorbox}
\usepackage{multirow}
\tcbuselibrary{breakable} 
\usepackage[table]{xcolor}  
\usepackage{array}          
\usepackage{enumitem} 
\usepackage{layouts} 

\newcommand{\ie}{\emph{i.e., }}
\newcommand{\eg}{\emph{e.g., }}

\newcommand{\ours}{{RWE-bench}}
\usepackage{pifont}

\icmltitlerunning{Can LLM Agents Deliver Evidence from Observational Research in Medical Databases?}

\begin{document}

\twocolumn[
  \icmltitle{Can LLM Agents Generate Real-World Evidence? \\ Evaluating Observational Studies in Medical Databases}



  \begin{icmlauthorlist}
    \icmlauthor{Dubai Li}{zju}
    \icmlauthor{Yuxiang He}{zju}
    \icmlauthor{Yan Hu}{zju}
    \icmlauthor{Yu Tian}{zju}
    \icmlauthor{Jingsong Li}{zju}
  \end{icmlauthorlist}

  \icmlaffiliation{zju}{Zhejiang University, Hangzhou, China}

  \icmlcorrespondingauthor{Yu Tian}{tyler@zju.edu.cn}
  \icmlcorrespondingauthor{Jingsong Li}{ljs@zju.edu.cn}

  \icmlkeywords{Large Language Models, Agents, Observational Studies, Real-World Evidence}

  \vskip 0.3in
]



\printAffiliationsAndNotice{}  

\begin{abstract}
  Observational studies can yield clinically actionable evidence at scale, but executing them on real-world databases is open-ended and requires coherent decisions across cohort construction, analysis, and reporting. Prior evaluations of LLM agents emphasize isolated steps or single answers, missing the integrity and internal structure of the resulting evidence bundle. To address this gap, we introduce \ours, a benchmark grounded in MIMIC-IV and derived from peer-reviewed observational studies. Each task provides the corresponding study protocol as the reference standard, requiring agents to execute experiments in a real database and iteratively generate tree-structured evidence bundles. We evaluate six LLMs (three open-source, three closed-source) under three agent scaffolds using both question-level correctness and end-to-end task metrics. Across 162 tasks, task success is low: the best agent reaches 39.9\%, and the best open-source model reaches 30.4\%. Agent scaffolds also matter substantially, causing over 30\% variation in performance metrics. Furthermore, we implement an automated cohort evaluation method to rapidly localize errors and identify agent failure modes. Overall, the results highlight persistent limitations in agents' ability to produce end-to-end evidence bundles, and efficient validation remains an important direction for future work. Code and data are available at \url{https://github.com/somewordstoolate/RWE-bench}.
\end{abstract}

\section{Introduction}

\begin{figure*}[t!]
    \centering
    \includegraphics[width=1\linewidth]{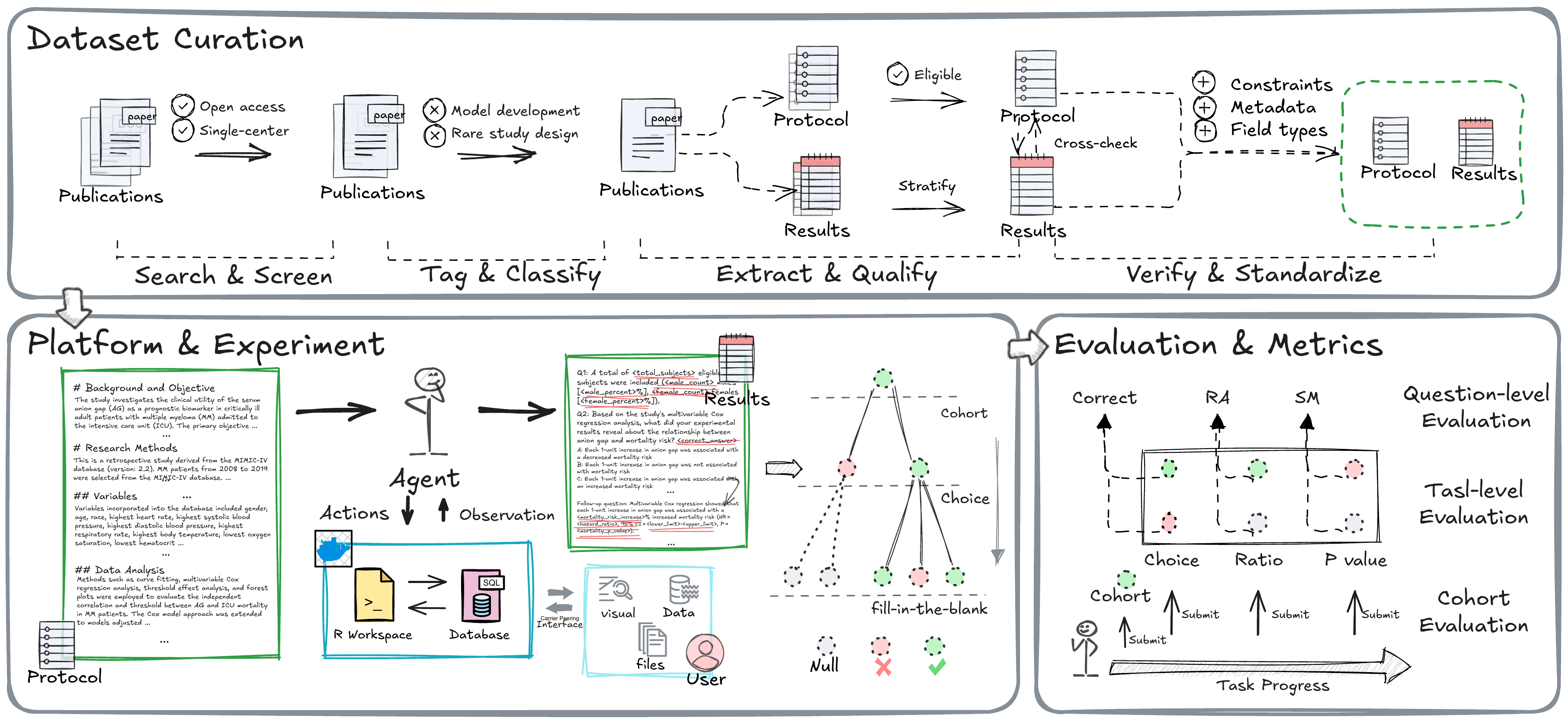}
    \caption{Overview of \textbf{\ours}\ construction, execution, and evaluation.
We curate a collection of peer-reviewed studies as benchmark tasks and require agents to reproduce the protocol-specified analyses on a real database.
Agents submit answers in a hierarchical format to form an evidence bundle.
We evaluate performance at both the question level and the task level, supplemented by an automatic verification of the generated cohorts.}
    \label{fig:rwe-bench overview}
\end{figure*}

Large language model (LLM) agents have recently shown the ability to execute multi-step tasks by combining planning, tool use, and iterative interaction with external environments, across domains such as computer use \citep{xieOSWorldBenchmarkingMultimodal2024,chezellesBrowserGymEcosystemWeb2025}, software engineering \citep{jimenezSWEbenchCanLanguage2023,wang2025openhands}, and deep research \citep{qiaoWebResearcherUnleashingUnbounded2025,liWebSailorNavigatingSuperhuman2025,liStreamliningEvidenceBased2025a}. These agent tasks share one or more favorable properties: they benefit from mature and openly accessible systems (\eg search engines), impose limited or relatively simple methodological constraints, and admit clearly defined end goals with established evaluation protocols. These characteristics substantially reduce ambiguity, allowing agents to work with high-quality context and feedback. However, real-world settings often present tasks with very different challenges.

In practice, a large fraction of societally consequential information is stored in closed, internally governed databases across enterprises, governments, and healthcare systems. These databases capture traces of real-world operations—transactions, service delivery, resource allocation, and longitudinal outcomes—and thus contain substantial latent value for measurement, causal analysis, and decision support. Yet turning such records into actionable evidence is often difficult: access is restricted, data structures are weakly documented, and extracting reliable conclusions requires a sequence of interdependent analytical operations.

Healthcare offers a particularly consequential instance of this challenge. Large-scale observational data such as electronic health records (EHRs) and claims databases can be transformed into real-world evidence (RWE), which informs clinical and policy decisions and is increasingly incorporated into regulatory and translational processes \citep{purpuraRoleRealWorldEvidence2022,dangRealworldEvidencePrimer2023}. However, generating RWE remains expensive and expert-intensive: observational studies require careful cohort construction, variable operationalization, and statistical modeling.
In practice, these steps typically demand close collaboration among clinicians, health informatics engineers, and data scientists.
The interdisciplinary nature of this process, together with the complexity of the data environment and the long-horizon analytical workflow, naturally raises the question of whether such studies can be automated by LLM-based agents.

Despite growing interest in medical research agents, existing evaluations often fall short of this setting. Many studies (i) assume preprocessed datasets or simplified schemas \citep{wangBioDSA1KBenchmarkingData2025,shiEHRAgentCodeEmpowers2024}, (ii) focus on isolated tasks that avoid long-horizon decision coupling \citep{wangMakingLargeLanguage2026,NEURIPS2024_99e81750}, or (iii) evaluate success primarily by final answers, overlooking the internal structure of evidence bundle \citep{wangBioDSA1KBenchmarkingData2025,chenScienceAgentBenchRigorousAssessment2024}. Consequently, it remains unclear whether current agents can conduct complete analyses on raw clinical databases in a manner that aligns with established research practice.

In this work, we study LLM agents for observational research in realistic settings. Our contributions include:
\begin{itemize}[noitemsep,topsep=0pt]
    \item \textbf{\ours:} a benchmark of 162 peer-reviewed observational studies within an end-to-end environment, featuring a hierarchical evaluation framework.
    \item \textbf{Large-scale evaluation on \ours}: an assessment of six open- and closed-source models across three scaffolds to quantify agent proficiency in generating complete clinical evidence.
    \item \textbf{Automated cohort verification:}  a rule-integrated pipeline for rapid cohort validation, utilizing LLM judges backed by a human-annotated evaluation set.
\end{itemize}


\section{\ours}
\label{benchmark}

Observational studies are inherently open-ended, holistic, and subject to uncertainty.
They require identifying appropriate target populations, selecting analysis strategies tailored to specific research questions, and iteratively modeling results based on intermediate findings. Owing to this openness, evaluating such tasks cannot rely on unit tests or held-out test sets with fixed ground-truth answers. To address this challenge, we introduce \textbf{\ours}, which uses peer-reviewed studies as reference standards, requiring agents to conduct experiments following the reported methods as protocols, and complete the resulting evidence bundle. To isolate the effect of the evidence structure itself on agent performance, evaluation points are organized in a tree-structured manner. \cref{fig:rwe-bench overview} illustrates the overall design of the benchmark.

\subsection{Task Description}
We formalize the agent’s task under a standard experimental framework.
The input is
\begin{equation}
    x = (P, D, \mathcal{Q}),
\end{equation}
where $P$ denotes the observational study protocol, $D$ the dataset. The question set $\mathcal{Q}$ is organized into two stages,
\begin{equation}
    \mathcal{Q} = \mathcal{Q}^{(0)} \cup \mathcal{Q}^{(1)},
\end{equation}
where $\mathcal{Q}^{(0)}$ consists of verification questions (\ie multiple-choice questions) that assess intermediate analytical results, and $\mathcal{Q}^{(1)}$ consists of follow-up fill-in-the-blank questions that are conditionally revealed.
Access to $\mathcal{Q}^{(1)}$ is gated by the correctness of the agent’s responses to $\mathcal{Q}^{(0)}$.

The agent interacts with the environment over multiple steps, performing concrete operations such as writing analysis code and submitting answers. In addition, before answering any questions, the agent must submit a table of the constructed study cohort for verification (see \cref{cohort_evaluation}).

After completing all interaction steps, the agent outputs a set of answers
\begin{equation}
    \hat{A} = \{\hat{a}_1, \ldots, \hat{a}_m\},
\end{equation}
where each $\hat{a}_i$ corresponds to a research question $q_i \in \mathcal{Q}$.

\subsection{Dataset Curation}
In our work, we adopt MIMIC-IV (v2.2) as the underlying database, as it is accessible, and curate benchmark materials from peer-reviewed studies conducted on this dataset. We use GPT-4.1 to accelerate certain data processing tasks, with manual checks performed to ensure accuracy.

According to the inclusion and exclusion criteria we defined, we selected 165 preliminary eligible open access single-center studies from 1,374 articles retrieved from PubMed (search strategy in \cref{fig:search_strategy}).
Most included studies follow STROBE\footnote{\href{https://www.strobe-statement.org}{STROBE} (STrengthening the Reporting of OBservational studies in Epidemiology) is a widely adopted international guideline for reporting observational studies.} and therefore adopt a consistent, well-structured reporting format. We leverage this structure to construct, for each \ours\ task, a \emph{protocol} from the full text and \emph{results} from the abstract. In addition, STROBE-style reporting typically separates methodological specification from outcome reporting, which helps limit result leakage during protocol construction.

Each protocol consists of three components: (i) \textit{background and objective}, (ii) \textit{research methods}, and (iii) \textit{questions}. The \textit{background and objective} are obtained via an LLM-based summarization of the Introduction section of each study. The \textit{research methods} are extracted directly from the Methods section. We then manually inspect the Methods content to reduce potential outcome leakage, removing or masking any text that could plausibly encode results. We also remove irrelevant details (\eg trial approval identifiers), supplement missing predefined elements not explicitly specified in the Methods (e.g., analytical formulas), and download and link supplementary materials to the corresponding tasks when they contain required information.

For results, the LLM is instructed to extract abstract sentences that contain numerical values reflecting experimental outcomes (instruction provided in \cref{fig:result_extraction_prompt}). For each task, we mask each extracted numerical value $a_i$ and replace it with a placeholder field, thereby converting the original sentence into a fill-in-the-blank query. Each blank (\ie each field) is treated as a question $q_i \in \mathcal{Q}^{(1)}$. All field values were type-checked to ensure they were numerical. 

To avoid information leakage, we introduce three LLM-based safeguards: (i) extracted sentences are categorized as \textit{baseline characteristics}, \textit{primary results}, or \textit{additional results}; \textit{additional results} are excluded because they often depend on \textit{primary results}; (ii) we instructed the LLM to answer masked fields based solely on the protocol context, performing this step twice and applying manual corrections to the protocol text where necessary; and (iii) for each \textit{primary result}, we prepend a single-choice question $q_j \in \mathcal{Q}^{(0)}$ that tests the reported qualitative phenomenon, and the corresponding fill-in-the-blank items can be attempted only if the correct option is selected. All single-choice questions are reviewed and manually revised when necessary, most commonly when options are overly specific or restrictive.

Following the \textit{ClinicalTrial.gov}\footnote{\href{https://clinicaltrials.gov/policy/results-definitions}{Results Data Element Definitions for Interventional and Observational Studies}}, each field was categorized into one of several standardized types: \textit{ratio}, \textit{confidence interval}, \textit{p-value}, \textit{count}, \textit{proportion}, \textit{AUC}, or \textit{numeric}. Ultimately, 162 tasks containing at least a \textit{ratio} or a \textit{p-value} were included in \ours. Details on benchmark construction and statistics are provided in \cref{app:dataset_curation} and \cref{app:dataset_statistics}.

\begin{figure}[ht]
  \begin{center}
    \centerline{\includegraphics[width=\columnwidth]{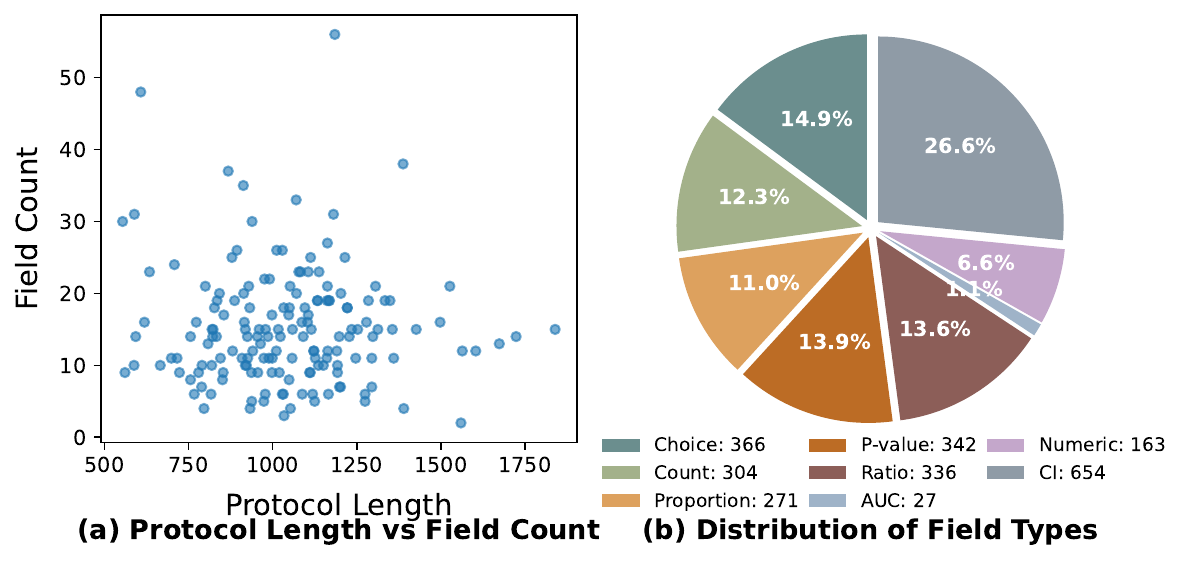}}
    \caption{
      Dataset statistics. (a) Protocol length (measured in word count) versus field count distribution. (b) Field type composition.
    }
    \label{fig:combined_scatter_pie_chart}
  \end{center}
  \vskip -0.1in
\end{figure}

\subsection{Environment Setup}

The agent research environment is designed to closely mirror widely adopted community practices, with the goal of reflecting how RWE studies are conducted in practice. 

\paragraph{Database}
Our benchmark is built on the MIMIC-IV (v2.2) \citep{johnsonMIMICIVFreelyAccessible2023}, which contains de-identified electronic health records from Beth Israel Deaconess Medical Center (BIDMC) for patients admitted to the emergency department or intensive care unit between 2008 and 2019.
The database comprises 299,712 patients, 431,231 hospital admissions, and 73,181 ICU stays.
We also construct derived tables using the mimic-code repository, which is built to enable reproducibility in critical care research \citep{johnson2018mimic}.
All data are stored in a PostgreSQL database.
We intentionally avoid task-specific customization beyond the standard MIMIC-IV setup.
This design choice ensures compatibility with other observational databases and enables the benchmark to support a broader range of similar tasks without environment-dependent assumptions.

\paragraph{Coding environment}
R is well suited for biomedical statistics and is widely used in observational research.
In our benchmark, most included studies adopt R, a choice further supported by prior work indicating that R-based analyses can be more agile than Python counterparts \citep{wangMakingLargeLanguage2026}.
Accordingly, we adopt R (v4.5.2) as the agent’s programming language and provide a standardized environment with commonly used packages for database access, data processing, and observational analysis (\eg propensity score modeling and causal inference).
A dedicated workspace was given to agents, which includes a R script for correctly establishing database connections and, when necessary, auxiliary files such as ICD code lists or predefined variables to facilitate faithful experiment execution.

\paragraph{Plug-and-Play}

Both the coding workspace and the database are provisioned as Docker containers that communicate over an isolated Docker network.
For each task execution, a dedicated database user is automatically created to enforce sandboxing and removed upon completion.
To support large-scale concurrent experiments, we integrate Ray and Joblib as parallel execution backends, together with randomized jitter-based retry mechanisms to mitigate conflicts arising from simultaneous database access.
Users can optionally retain execution traces for reproducibility and auditing.
We additionally provide a lightweight monitoring tool to visualize the agent’s actions and execution progress in real time.

\subsection{Evaluation and Metrics}

Most prior data science benchmarks evaluate systems solely on final outputs. While this may be adequate for simple or single-step tasks, it breaks down for complex, multi-step workflows such as biomedical data analysis, where intermediate conclusions directly condition how the final result should be interpreted. For example, a study may report a positive headline association, yet an intermediate validity check—such as a negative-control outcome—exhibits a comparably strong “effect,” reframing the finding from a plausible causal signal to evidence of residual confounding. This motivates evaluation at multiple granularities and across dimensions. We therefore structure our evaluation at two levels.

\subsubsection{Semantic Alignment}

We introduce question-level evaluation metrics that focus on the semantic alignment between each agent's answer and the corresponding result reported in the original studies.

According to \textit{ClinicalTrial.gov}\footnote{\href{https://cdn.clinicaltrials.gov/documents/data-prep-checklist-om-sa.pdf}{Outcome Measure Data Preparation Checklist}}, \textit{statistical test of hypothesis} and \textit{effect estimate} belong to \textit{statistical analysis} section. Accordingly, we apply semantic alignment evaluation metrics for both of them. 

For \textit{effect estimates}, it corresponds to a pair of \textit{ratio} field and \textit{confidence interval} fields in our settings. Since all included studies report relative effects, we adopt the \textbf{Regulatory Alignment Rate (RAR)} to assess alignment between agent outputs and publications, which regulatory alignment is an epidemiological binary metrics \citet{franklinEmulatingRandomizedClinical2021,franklinNonrandomizedRealWorldEvidence2020}: The agent reproduces the \emph{direction} and \emph{statistical significance} of the reported effect.
\begin{equation}
\text{RA} = \mathbb{I}\Big[
\begin{aligned}
&(\hat{r}-1)(r-1) > 0 \\
&\land\; (1 \in [\hat{l},\hat{u}]) \Leftrightarrow (1 \in [l,u])
\end{aligned}
\Big].
\end{equation}
Here, $\hat{r}$ and $r$ denote the effect estimates from the agent and the publication, respectively, $[\hat{l}, \hat{u}]$ and $[l, u]$ the corresponding confidence intervals, and $\mathbb{I}(\cdot)$ the indicator function.

For \textit{statistical tests of hypotheses}, the relevant field is the \textit{p-value}.
We assess whether the agent and the publication reach the same conclusion on statistical significance under a conventional threshold.
Specifically, a \textbf{Significance Match (SM)} is recorded if both p-values are either below or above the significance level $\alpha = 0.05$:
\begin{equation}
\text{SM} =
\mathbb{I}\!\left[
(\hat{p} < 0.05) \Leftrightarrow (p < 0.05)
\right],
\end{equation}
Here, $\hat{p}$ and $p$ denote the agent-generated and reported p-values, respectively. We then report the \textbf{Significance Match Rate (SMR)} as the macro-average of SM across tasks.

For the remaining fields, which primarily reflect descriptive statistics, effective metrics for direct assessment are often lacking.
We therefore evaluate part of them (belonging to primary results) using multiple-choice accuracy \textbf{ACC}, which can be treated as a simplified semantic evaluation.


\subsubsection{Task-level Metrics}

In our benchmark, each task consists of a sequence of interdependent questions whose answers jointly form an evidence bundle supporting a specific quantitative conclusion.
As a result, task-level assessment is required to capture the overall quality and completeness of the generated evidence.

We first introduce the \textbf{Success Rate (SR)}, where a task is considered successful if all multiple-choice questions are answered correctly and all fields related to \textit{statistical analysis} achieve semantic alignment with the corresponding human-reported results.

Beyond correctness, we further evaluate efficiency and task completion behavior.
Specifically, we report the average number of interaction steps (\textbf{Steps}) taken by the agent to complete a task, and the \textbf{Completion Rate (CR)}, which measures whether the agent successfully answers all visible questions within the predefined resource constraints.
These metrics characterize the dynamics of agent execution and help differentiate the behaviors of different agent configures.

\section{EXPERIMENTS}
\label{experiment}

\subsection{Experiment Setup}

\begin{table*}[ht]
\caption{Main results on \ours.
Metrics are reported as macro-averages, with boldface denoting the best-performing configuration.
All metrics are expressed as percentages (\%) except for Steps.
In total, the benchmark includes 162 tasks, covering 366 \textit{choice}, 313 \textit{ratio} with CIs, and 342 \textit{p-value} fields.}

\label{tab:main_results}
\centering
\resizebox{0.8\textwidth}{!}{%
\begin{tabular}{@{}lcccccc@{}}
\toprule
\multicolumn{1}{c}{}                                 & \multicolumn{3}{c}{\textbf{Question-level Metrics}}             & \multicolumn{1}{l}{}                        & \textbf{Task-level Metrics} & \multicolumn{1}{l}{} \\ \cmidrule(lr){2-4}\cmidrule(lr){5-7} 
\multicolumn{1}{c}{\multirow{-2}{*}{\textbf{Model}}} & \textbf{ACC}        & \textbf{RAR}         & \textbf{SMR}         & \cellcolor[HTML]{EFEFEF}\textbf{SR}         & \textbf{Steps}              & \textbf{CR}          \\ \midrule
\multicolumn{7}{c}{\textbf{MLAB}} \\ \midrule
GPT-4.1 & 31.5 ± 1.0 & 16.5 ± 1.8 & 20.4 ± 1.0 & \cellcolor[HTML]{EFEFEF}10.3 ± 0.7 & 79.5 ± 1.8 & 33.1 ± 2.3 \\
Claude-Sonnet-4 & 60.9 ± 1.2 & 50.1 ± 1.8 & 46.8 ± 2.1 & \cellcolor[HTML]{EFEFEF}31.1 ± 2.5 & 49.5 ± 1.5 & 82.1 ± 0.9 \\ \midrule
\multicolumn{7}{c}{\textbf{OpenHands}} \\ \midrule
MiniMax-M2.1 & 59.8 ± 0.1 & 47.5 ± 2.7 & 47.6 ± 2.6 & \cellcolor[HTML]{EFEFEF}29.4 ± 1.9 & 53.7 ± 1.4 & 91.4 ± 1.8 \\
GPT-4.1 & 3.6 ± 0.6 & 1.6 ± 0.4 & 1.8 ± 0.2 & \cellcolor[HTML]{EFEFEF}1.2 ± 0.4 & 49.6 ± 1.1 & 65.4 ± 1.9 \\
Claude-Sonnet-4 & 63.7 ± 1.7 & 53.4 ± 0.8 & 53.3 ± 1.9 & \cellcolor[HTML]{EFEFEF}35.0 ± 1.7 & 45.9 ± 0.2 & 96.9 ± 0.7 \\ \midrule
\multicolumn{7}{c}{\textbf{RWEAgent}} \\ \midrule
Qwen3-30B-A3B & 18.8 ± 3.1 & 10.9 ± 2.8 & 11.5 ± 3.0 & \cellcolor[HTML]{EFEFEF}6.2 ± 3.1 & 43.2 ± 2.0 & 31.9 ± 1.5 \\
MiniMax-M2.1 & 61.9 ± 2.0 & 49.3 ± 1.7 & 47.0 ± 4.0 & \cellcolor[HTML]{EFEFEF}30.4 ± 0.8 & 44.4 ± 0.5 & 95.1 ± 1.4 \\
GLM-4.7 & 62.0 ± 0.5 & 48.9 ± 2.4 & 46.6 ± 0.3 & \cellcolor[HTML]{EFEFEF}29.6 ± 0.7 & 48.1 ± 1.4 & 93.6 ± 1.8 \\
GPT-4.1 & 62.9 ± 1.5 & 47.0 ± 3.4 & 45.4 ± 2.4 & \cellcolor[HTML]{EFEFEF}30.9 ± 1.9 & 27.9 ± 0.6 & 88.1 ± 1.8 \\
O4-mini & 66.4 ± 0.6 & 51.4 ± 0.8 & 48.0 ± 0.7 & \cellcolor[HTML]{EFEFEF}35.2 ± 1.6 & 24.1 ± 0.8 & 80.5 ± 1.1 \\
\textbf{Claude-Sonnet-4} & \textbf{71.2 ± 1.5} & \textbf{60.3 ± 0.9} & \textbf{55.7 ± 1.4} & \cellcolor[HTML]{EFEFEF}\textbf{39.9 ± 1.5} & 35.7 ± 0.9 & 99.4 ± 0.0 \\ \bottomrule
\end{tabular}%
}
\end{table*}

\paragraph{Agent scaffolds}
We evaluate three agent scaffolds. 
\textbf{MLAB} \citep{huangMLAgentBenchEvaluatingLanguage2024} follows a ReAct-style design, decomposing reasoning into distinct stages including \textit{Reflection, Research Plan and Status, Fact Check, Thought, and Action}.
\textbf{OpenHands} \citep{wang2025openhands} is a general-purpose coding agent built upon the CodeAct framework, featuring relatively mature mechanisms for planning, memory management, and task execution.
\textbf{RWEAgent} is adapted from the EHRAgent architecture \citep{shiEHRAgentCodeEmpowers2024}, motivated by limitations in long-horizon execution and action compatibility observed in our setting.
Further details are provided in Appendix~\ref{app:agent_scaffolds}.

\paragraph{Models}
We evaluate close-source models with different positioning, including Claude Sonnet-4 \citep{anthropicIntroducingClaude42025}, OpenAI O4-mini\citep{openaiIntroducingOpenAIO32025}, and OpenAI GPT-4.1\citep{openaiIntroducingGPT41API2025}. For open-source models, we benchmark ZhiPu GLM-4.7 (358B)\citep{5team2025glm45agenticreasoningcoding}, MiniMax M2.1 (229B)\citep{minimaxMiniMaxM21Significantly2025}, and Qwen3-30B-A3B\citep{qwen3technicalreport}, selected based on their parameter scale. We also explored models fine-tuned for the medical field, but most of them failed to effectively perform long-horizon coding tasks and are thus excluded from evaluation. We invoke all models via APIs and fix the temperature at 0.6 when supported. Since O4-mini does not expose a temperature setting, we instead set its reasoning effort to medium.

\paragraph{Implementation details}
To avoid unnecessary consumption of database and computational resources, the agents are configured with a maximum of 100 steps, a time limit of 20 minutes per step, and an overall runtime cap of 2 hours. 
To prevent trial-and-error behavior, each question permits only a single submission.
Each task is run three times, and we report the mean and one standard error.

\subsection{Main Experiment}
\label{main_experiment}

\paragraph{Overall performance}

\cref{tab:main_results} summarizes the main experimental results.
Overall, only a small subset of models (\ie O4-mini, Claude-Sonnet-4) achieves an average score above 50\% on question-level metrics.
Among these metrics, ACC is consistently higher than RA and SM, reflecting the effect of the multiple-choice gating mechanism.
However, when evaluated at the evidence-bundle level, even the best-performing agent, RWEAgent (Claude-Sonnet-4), attains a SR of only 39.9\%. This gap highlights the substantial difference between reproducing individual analytical outcomes and constructing a fully aligned evidence bundle that matches human-reported results.

In addition, we find that RWEAgent (Qwen3-30B-A3B), MLAB (GPT-4.1), and OpenHands (GPT-4.1) achieve the lowest SR and CR. A closer inspection of the incomplete runs suggests two distinct failure modes. OpenHands (GPT-4.1) often terminates prematurely: after submitting the first answer, it explicitly asks whether it should continue with subsequent questions and then ends the execution, a behavior that is rarely observed in other agents. In contrast, failures of RWEAgent (Qwen3-30B-A3B) and MLAB (GPT-4.1) primarily reflect capability limitations—these agents either stop early due to unresolvable issues or exhaust the allowed attempt budget before completing the full task (see \cref{app:incomplete_tasks}).

\paragraph{Impact of agent scaffolds.}
We find that the choice of agent scaffold has a substantial impact on task performance, with effects that vary across models.
For instance, GPT-4.1 attains an SR above 30\% under RWEAgent, but drops to about 10\% under MLAB and to roughly 1\% under OpenHands; in contrast, Claude-Sonnet-4 and MiniMax-M2.1 exhibit a smaller yet consistent decline across the same agent frameworks.

Across all evaluated models, RWEAgent consistently yields the strongest performance, and all closed-source models achieve higher SRs than open-source counterparts under this scaffold.
In contrast, MLAB and OpenHands exhibit higher average step counts and lower CRs, indicating less efficient exploration and greater difficulty in completing all required subtasks within the imposed resource constraints. 

Different scaffolds also exhibit distinct token consumption patterns. As shown in \cref{fig:token_cost_avg}, OpenHands incurs significantly higher usage than other scaffolds across the same models. Specifically, MiniMax-M2.1 records the highest average token consumption under this framework, exceeding 1.4 million tokens per task, yet we do not observe a corresponding improvement in performance.

All evaluated scaffolds are general-purpose and not optimized for observational research.
This suggests that specialized agent designs with improved long-horizon context engineering may further enhance efficiency and reliability in automated real-world evidence generation.

\paragraph{Open-source vs. closed-source models.}
Two strong open-source models, MiniMax-M2.1 and GLM-4.7, are evaluated based on their competitive standing on the SWE-bench leaderboard and performance comparable to Claude-Sonnet-4 \cite{jimenezSWEbenchCanLanguage2023}.
Despite their strong capabilities, these top-tier open-source models still underperform relative to closed-source counterparts on observational research tasks.
Nevertheless, the performance gap is not too large, and continued progress in open-source models is likely.
This trend is particularly encouraging for applications involving large-scale private data, where open-source solutions are often preferred or required.

We compare the open-source reasoning model GLM-4.7 with the closed-source reasoning model O4-mini. Under the same agent scaffold, we observe pronounced differences in their behavioral patterns, including the interaction step number (see \cref{fig:code_action_stat}) and token consumption (see \cref{fig:token_cost_avg}).

We further examine Qwen3-30B-A3B, which has the smallest parameter scale among the evaluated models and is representative of commonly used fine-tuned models in the 7B–32B range. We also experimented with other medical models of comparable size, including MedCopilot-14B \cite{xuMedAgentGymTrainingLLM2025a}, which is renowned for its training on medical coding tasks. We found that most of these models failed to maintain effective instruction following or sustained reasoning under long-context settings. As a result, completing more complex long-horizon medical programming tasks may require models with larger parameter scales.

\begin{figure}[h]
  \begin{center}
    \centerline{\includegraphics[width=\columnwidth]{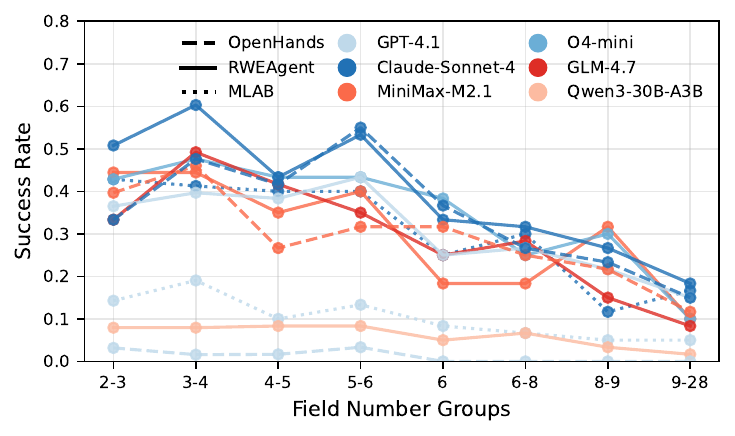}}
    \caption{
      Success rates across task groups stratified by field count. The x-axis groups tasks by the number of fields, with labels indicating the corresponding intervals.
    }
    \label{fig:sr_across_field_groups}
  \end{center}
\vskip -0.2in
\end{figure}

\paragraph{Effect of field cardinality.}
Tasks in \ours\ vary in the number of fields, leading to differences in intrinsic difficulty. To analyze this effect, we group tasks according to the total number of fields (including \textit{choice}, \textit{ratio}, and \textit{p-value}), partitioning them into eight bins ordered from small to large, each containing 20–21 tasks.
The SR values across task groups are reported in \cref{fig:sr_across_field_groups}.

As expected, most models exhibit a clear downward trend in SR as the number of required fields increases, suggesting that tasks become more challenging with higher field cardinality. Notably, stronger agents tend to show steeper declines: they substantially outperform weaker agents on tasks with fewer required fields, leading to a more pronounced stratification between agent configurations. Motivated by this pattern, we construct \ours-hard by selecting tasks with at least nine required fields, thereby increasing overall task difficulty (the highest SR reaches 23\%) while reducing evaluation cost (see \cref{app:rwe-bench-hard}).

\subsection{Cohort Evaluation}
\label{cohort_evaluation}

Observational studies differ fundamentally from software or model development: distinct analytical operations can yield similar results, while the underlying statistical meaning and causal interpretation may diverge substantially. Although a tree-structured evidence bundle can partially expose internal inconsistencies, cohort construction is an fundamental step that can systematically bias all subsequent analyses; it therefore benefits from quality checks.

The conventional gold-standard approach of cohort evaluation—expert manual chart review of sampled cases---is accurate but expensive and difficult to scale. Recent work has explored using LLMs for case adjudication in observational research and demonstrated promising potential \cite{schuemieStandardizedPatientProfile2025}. Building on this line of work, we employ an automated cohort evaluation approach that combines heuristic checks with an LLM-as-a-judge to provide rapid, preliminary screening of the agent-constructed cohorts.

\paragraph{\textsc{CohortEval}}
To assess the reliability of LLM-based cohort adjudication, we develop \textsc{CohortEval}, a dedicated test set comprising 10 tasks sampled from \ours. Each task consists of a cohort of 140 cases, which includes a subset of the cohort originally constructed by the agent, supplemented by counterfactual samples generated for class balance and adversarial testing. The ground truth for these samples was established through annotation by two medical graduate students. More details can be found in \cref{app:cohorteval_details}.

We build the LLM cohort judge using the prompt in \cref{fig:cohort_judge_prompt}. Given an individual patient case and the protocol eligibility criteria, the judge is instructed to identify attributes that violate inclusion/exclusion rules and decide whether the case should be excluded. We evaluate multiple judge models on \textsc{CohortEval} by comparing their decisions against the gold standard, reporting sensitivity, specificity, PPV, and NPV in \cref{tab:cohorteval_main_results}. We select GPT-5 as the default cohort judge for evaluating agent-constructed cohorts.

\begin{table}[h]
\caption{Performance on \textsc{CohortEval} averaged over three runs (mean $\pm$ SE \%). * denotes models fine-tuned on medical corpora.}
\centering
\resizebox{\linewidth}{!}{
\begin{tabular}{@{}lcccc@{}}
\hline
\textbf{Model}  & \textbf{Sens} & \textbf{Spec} & \textbf{PPV} & \textbf{NPV} \\ \hline
GPT-5           & \textbf{93.1±0.4}      & 90.7±0.1      & 95.4±0.1     & \textbf{90.6±0.2}     \\
O4-mini         & 81.1±0.3             & 93.0±0.7        & \textbf{96.1±0.3}            & 78.0±0.2            \\
GPT-4.1         & 80.7±0.7      & \textbf{93.3±0.2}      & 88.2±1.8     & 79.1±0.5     \\
Qwen3-32B       & 69.2±0.3             & 92.3±0.7             & 94.5±0.5            & 66.5±0.1            \\
Lingshu-32B*     & 67.0±1.0             & 84.8±0.9             & 88.2±0.8            & 64.0±0.7            \\ 
ReasonMed-8B* & 78.5±0.0            & 76.1±0.0              & 87.5±0.0            & 66.9±0.0            \\  \hline

\end{tabular}%
}
\label{tab:cohorteval_main_results}
\end{table}
\paragraph{Cohort screening}

During each \ours\ task execution, the agent is required to export a cohort audit file before submitting its first answer: a table containing basic information for all included patients. This file must (i) include identifiers that uniquely determine the hospitalization/ICU admission records and (ii) provide the key variables needed to evaluate the protocol’s inclusion and exclusion criteria (\eg diagnosis codes, demographics). 

For cohort evaluation, we collect the cohort tables from one run among the three repeated executions for each task. Due to budget constraints, this screening is limited to cohorts generated by RWEAgent. We first apply a rule-based screening to filter out clearly invalid outputs: the table must be non-empty, every \texttt{subject\_id} must be retrievable in MIMIC-IV, and the cohort size must not exceed 50{,}000. For cohorts that pass these checks, we uniformly sample up to 100 patient cases and evaluate each case independently using the cohort judge. This comprehensive evaluation consumed approximately 200 million tokens. We define the exclusion rate as the fraction of sampled cases rejected by the judge; if this rate exceeds a threshold $C$ (set to $0.8$ in our experiments), the cohort is marked as unqualified.

\begin{figure}[ht]
  \begin{center}
    \centerline{\includegraphics[width=\columnwidth]{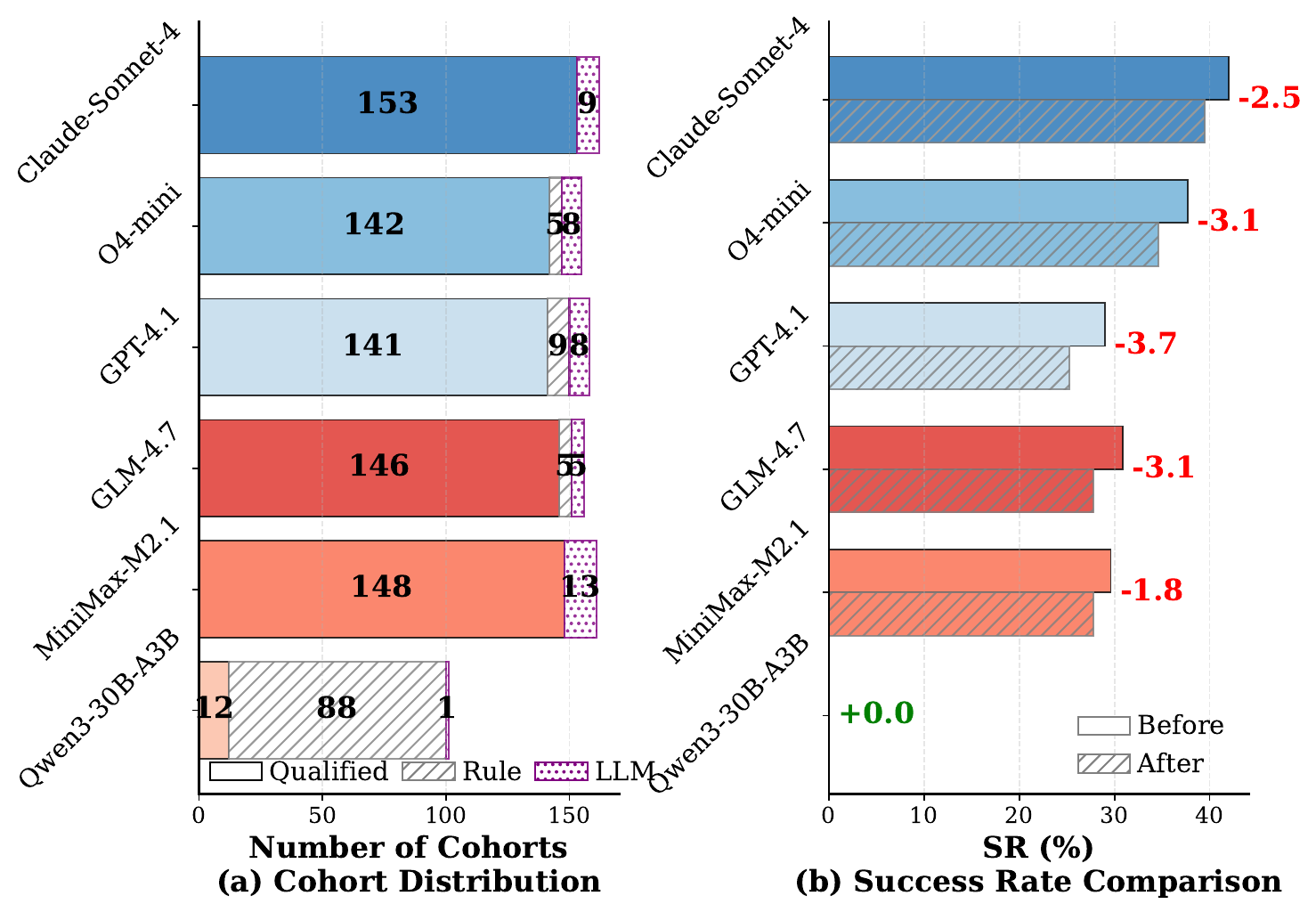}}
    \caption{
      Screening results of RWEAgent cohorts. (a) Counts of qualified cohorts versus cohorts filtered out by rule-based checks or by the LLM judge. (b) Change in SR after cohort screening.
    }
    \label{fig:cohort_screening_results}
  \end{center}
  \vskip -0.2in
\end{figure}

\cref{fig:cohort_screening_results} summarizes the screening results. Notably, Qwen3-30B-A3B generated a significant number of invalid cohorts (88 in total), including 27 cases containing fabricated \texttt{subject\_id}s. We conducted a manual review of the screening results to identify the causes of invalidity, such as ambiguous cohort definitions 
(see \cref{app:invalid_cohorts} for a detailed analysis). Regarding performance impact, most agents exhibited a slight decrease in SR of approximately 2--3\% after screening. Overall, LLM-based screening demonstrates potential utility in assisting humans with rapid error identification. This overhead motivates future research into specialized models and cost-effective systems.

\section{Related Work}
\label{related_work}

\paragraph{LLM Agent evaluation in data science.}
Data science has become a popular testbed for agent-based research automation\citep{guoDSagentAutomatedData2024,hongDataInterpreterLLM2024}. Existing data science benchmarks either emphasize final outcomes (e.g., \citet{huangMLAgentBenchEvaluatingLanguage2024}, \citet{chanMLEbenchEvaluatingMachine2024}) or code execution correctness (e.g., \citet{laiDS1000NaturalReliable2023,chenScienceAgentBenchRigorousAssessment2024,jingDSBenchHowFar2025}). While effective for narrow tasks, these designs are largely single-target and result-oriented, overlooking the multi-step, contextual nature of real-world analysis. Recent efforts such as MLR-bench \citep{chenMLRbenchEvaluatingAI2025} and DSEval \citep{bediHolisticEvaluationLarge2026} explore process-level evaluation, but remain limited in reliability or expressiveness. DAgent \citep{xuDAgentRelationalDatabasedriven2025} enables agents to query relational databases for report generation, yet it lacks sufficient depth in statistical computation.
Biomedical data science benchmarks such as BioCoder \citep{tangBioCoderBenchmarkBioinformatics2024} and BioDSbench \citep{wangMakingLargeLanguage2026} further simplify evaluation by focusing on code validity, while BioDSA-1k \citep{wangBioDSA1KBenchmarkingData2025} relies on preprocessed datasets, which risks overstating agent's practical ability. In contrast, \ours\ encodes soft constraints in the form of protocols and sets up multi-step, hierarchical, and fine-grained questions to simulate an end-to-end real-world evidence generation.

\paragraph{LLM evaluation in medical domains.}
Medical benchmarks span diverse tasks. Some focus on relatively self-contained problems such as note generation \citep{yimAcibenchNovelAmbient2023,mtsamplesTranscribedMedicalTranscription2023} or clinical QA \citep{flemingMedAlignClinicianGeneratedDataset2024,vilaresHEADQAHealthcareDataset2019}, which require little environmental design.
While such isolation facilitates targeted model optimization, its relevance to more complex clinical settings is unclear. To address this limitation, subsequent benchmarks can be categorized as two complementary directions.
One direction broadens evaluation scope by incorporating diverse medical tasks, enabling multi-dimensional assessment across heterogeneous capabilities.
For example, MedHELM \citep{bediHolisticEvaluationLarge2026} integrates a wide range of real-world healthcare tasks to assess cross-domain generalization, while MedAgentGym \citep{xuMedAgentGymTrainingLLM2025a} provides a unified platform for training and evaluating agents on medical coding tasks.
The other emphasizes realism by moving evaluation closer to real-world infrastructures, such as MedAgentBench \citep{jiangMedAgentBenchRealisticVirtual2025}, which builds on a FHIR-based medical database.
Although these settings introduce complexity, they offer closer alignment with deployment scenarios.
Building on these directions, \ours\ focuses on evaluating agents on structurally complex research tasks that comprise multiple interdependent steps within a realistic research infrastructure.

We summarize the key characteristics of these related benchmarks in \cref{tab:benchmark_comparison}.

\section{Discussion}

\paragraph{Open Benchmarking as a Catalyst}

Constructing an RWE benchmark requires an observational database that is accessible, supports a substantial body of validated clinical findings, and is accompanied by necessary methodological detail to enable reproducible analysis. Under such stringent criteria, MIMIC emerges as one of the few viable candidates. However, it is crucial to acknowledge a potential limitation: the prevalence of MIMIC database in LLM pre-training corpora may lead to data contamination, potentially inflating model performance. Nevertheless, our observations indicate that distinct performance gaps remain among models when addressing more complex tasks, with overall proficiency remaining suboptimal. Consequently, we position \ours\ as a centralized hub for community-driven development, providing a shared foundation to refine agentic capabilities before validating their generalizability on proprietary databases.

\paragraph{Scalable Pre-validation Screening}
As automated systems accelerate the generation of clinical evidence, efficient screening becomes essential to prioritize resources before committing to downstream validation. 
While benchmarks like ScienceAgentBench \cite{chenScienceAgentBenchRigorousAssessment2024} and Paperbench \cite{starace2025paperbench} integrate LLMs into evaluation, they remain bottlenecked by the high manual overhead of designing and maintaining stage-wise rubrics. 
In this work, we introduce a lightweight, task-agnostic LLM-assisted screening layer. 
Rather than replacing human review, this system functions as an early-warning mechanism to flag suspicious artifacts, thereby streamlining the validation pipeline. 
However, limitations persist: this approach does not fully resolve the challenge of assessing cohort completeness (recall), and the screening reliability may be compromised by agent-induced hallucinations or information omissions. 
Achieving fully automated, robust, and cost-effective screening remains an open research question, necessitating the co-evolution of AI capabilities and evaluation frameworks.

\section{Conclusion}
We introduce \ours, a benchmark designed to evaluate LLM agents on observational studies.
We define task formats based on common medical practices, using protocols to standardize operations and constrain open-ended research. A layered evidence structure is built to minimize bias affecting the Agent's behavior. This structure aligns with real-world research, facilitating validation set expansion and transition to practical scenarios. Our evaluation reveals a 10\% SR gap between MiniMax-M2.1 and Claude-Sonnet-4 despite similar coding proficiency, while also highlighting the critical potential of domain-tailored scaffolds. Given the substantial headroom for improvement in current capabilities, \ours\ establishes a foundational testbed to guide the evolution of automated RWE generation, demonstrating a scalable paradigm and infrastructure that can be adapted to mine insights from private databases in various contexts.




\section*{Impact Statement}

This work aims to advance the field of machine learning by studying the capabilities and limitations of LLM agents in conducting complex observational research and real-world evidence analyses. By providing a structured evaluation framework and benchmarked tasks, our study contributes to a better understanding of how such agents perform in realistic, open-ended research environments, with potential benefits for reproducibility, efficiency, and transparency in data-driven scientific workflows.

From a societal and ethical perspective, this work does not introduce new machine learning models or deploy systems intended for direct clinical or policy decision-making. Instead, it focuses on methodological evaluation and analysis. Nevertheless, the findings may indirectly influence future applications of LLM-based agents in sensitive domains such as healthcare research. Misinterpretation or overreliance on automated agents for observational studies without appropriate human oversight could lead to flawed analyses or misleading conclusions. Our work explicitly highlights these limitations and emphasizes the necessity of rigorous protocols, auditability, and human-in-the-loop validation.

We do not anticipate immediate negative societal consequences arising from this work. We hope that, by clarifying both the strengths and failure modes of LLM agents in real-world research settings, this study will support the responsible development and deployment of machine learning systems in scientific and medical research.


\bibliography{rwe-bench}
\bibliographystyle{icml2026}

\newpage
\appendix
\onecolumn


\section{Data \& Code Availability}
To facilitate reproducibility, we will release the code and data to the public upon acceptance.

\section{Related Works}

\begin{table}[h]
\centering
\caption{Comparison of related benchmarks. \textbf{Input Scope} denotes the complexity of the task description (e.g., full paper vs. simple instruction). \textbf{Solution Granularity} indicates the scale of the generated code. \textbf{Domain Abbreviations}: Admin = Administration; DA = Data Analysis; ML = Machine Learning.}
\label{tab:benchmark_comparison}
\resizebox{\columnwidth}{!}{%
\begin{tabular}{@{}llllllll@{}}
\toprule
\textbf{Benchmark} & \textbf{Domain} & \textbf{Language} & \textbf{Source} & \textbf{Input Scope} & \textbf{Solution Granularity} & \textbf{Primary Verification} & \textbf{Size} \\ \midrule
EHRSQL\cite{leeEHRSQLPracticalTexttoSQL2022} & Med Admin & SQL & Questionnaire & Instruction & Sentence-level & Code Test & 24,405 \\
MedAgentBench\cite{xuMedAgentGymTrainingLLM2025a} & Med Admin & - & Clinician & Instruction & Sentence-level & Result Match & 300 \\
MLE-bench\cite{chanMLEbenchEvaluatingMachine2024} & ML & Python & Kaggle & Protocol & Repo-level & Code Test & 75 \\
Paperbench\cite{starace2025paperbench} & ML & Python & Publications & Paper & Repo-level & Rubric & 20 \\
DSBench\cite{jingDSBenchHowFar2025} & DA/ML & Python & Kaggle/ModelOff & Instruction & Script-level & Choice/Fill & 540 \\
ScienceAgentBench\cite{chenScienceAgentBenchRigorousAssessment2024} & DA/ML & Python & Publications & Instruction & Script-level & Execution & 102 \\
BioCoder\cite{tangBioCoderBenchmarkBioinformatics2024} & Biomed DA & Py/Java & GitHub & Instruction & Function-level & Code Test & 2,522 \\
BioDSbench\cite{wangMakingLargeLanguage2026} & Biomed DA & Py/R & Publications & Instruction & Script-level & Code Test & 293 \\
BioDSA-1k\cite{wangBioDSA1KBenchmarkingData2025} & Biomed DA & Python & Publications & Instruction & Script-level & Choice & 1,029 \\ \midrule
\ours\ (Ours) & Med DA & SQL+R & Publications & Protocol & Repo-level & Choice + Fill & 162 \\ \bottomrule
\end{tabular}%
}
\label{tab:related_works}
\end{table}

\section{Dataset Curation Details}
\label{app:dataset_curation}

\cref{fig:search_strategy} illustrates the search strategy employed in PubMed, which initially identified 1,374 records. These records were subsequently screened based on the following eligibility criteria:

\begin{enumerate}
    \item \textbf{Access:} Only Open Access publications were included.
    \item \textbf{Data Source:} Studies must utilize the MIMIC-IV (v2.2) dataset. We verified the specific version usage by employing a LLM to extract relevant text segments, followed by manual confirmation to ensure the study relied on version 2.2.
    \item \textbf{Study Objective:} We excluded review articles and technical papers focused solely on model development. We retained only studies aimed at generating real-world evidence.
    \item \textbf{Study Design:} In accordance with the STROBE guidelines, we restricted our selection to standard observational designs, specifically cohort, cross-sectional, and case-control studies.
\end{enumerate}

Following the classification and screening process, 165 studies advanced to the dataset construction phase. For each study, we employed a LLM—using the prompt detailed in \cref{fig:result_extraction_prompt}—to extract numerical data fields from the abstract. The original sentences corresponding to these fields were subsequently categorized into three types: \textit{baseline\_characteristics}, \textit{primary\_results}, and \textit{additional\_results}. The \textit{additional\_results} category typically encompasses subgroup or sensitivity analyses that expand upon the findings of the \textit{primary\_results}. To prevent data leakage, results classified as \textit{additional\_results} were excluded from the benchmark.

Next, we instructed the LLM to construct a multiple-choice question (MCQ) for each result identified as \textit{primary\_results}. The generation constraints required the MCQ to feature mutually exclusive options with a single unique correct answer corresponding to the primary result, ensuring that no information regarding the answer was leaked through the question stem or incorrect options.

Concurrently, we prompted the LLM (see \cref{fig:field_type_prompt_1}) to determine the specific data type for each extracted field, aiming to identify the definitive statistical analysis outcome of the RWE. We retained only those studies containing at least one valid statistical result field; consequently, 3 studies were excluded from the final dataset. All generated content underwent manual verification and refinement.

\begin{figure}[ht]
\vskip 0.2in
  \begin{tcolorbox}[colback=white,colframe=blue!75!black, width=\textwidth, title=Search Strategy]
    \begin{lstlisting}[basicstyle=\tiny\ttfamily, breaklines=true]
    ("MIMIC-IV"[Title/Abstract] OR "MIMIC IV"[Title/Abstract] OR mimiciv) 
    AND
    (cohort[Title/Abstract] OR "cohort study"[Title/Abstract] OR retrospective[Title/Abstract] OR prospective[Title/Abstract] OR observational[Title/Abstract] OR "observational study"[Title/Abstract] OR "case-control"[Title/Abstract] OR "cross-sectional"[Title/Abstract] OR "comparative study"[Title/Abstract] OR registry[Title/Abstract] OR "real-world"[Title/Abstract])
    NOT
    ("case reports"[Publication Type] OR "review"[Publication Type] OR "systematic review"[Publication Type] 
    OR "meta-analysis"[Publication Type] OR "editorial"[Publication Type] OR "letter"[Publication Type])
    \end{lstlisting}
  \end{tcolorbox}
    \cprotect
  \caption{Search strategy. It was performed on PubMed with a search date prior to August 19, 2025.}
  \label{fig:search_strategy}
\end{figure}

\section{Dataset statistics}
\label{app:dataset_statistics}

\begin{figure}[!h]
  \vskip 0.2in
  \begin{center}
    \centerline{\includegraphics[width=\columnwidth]{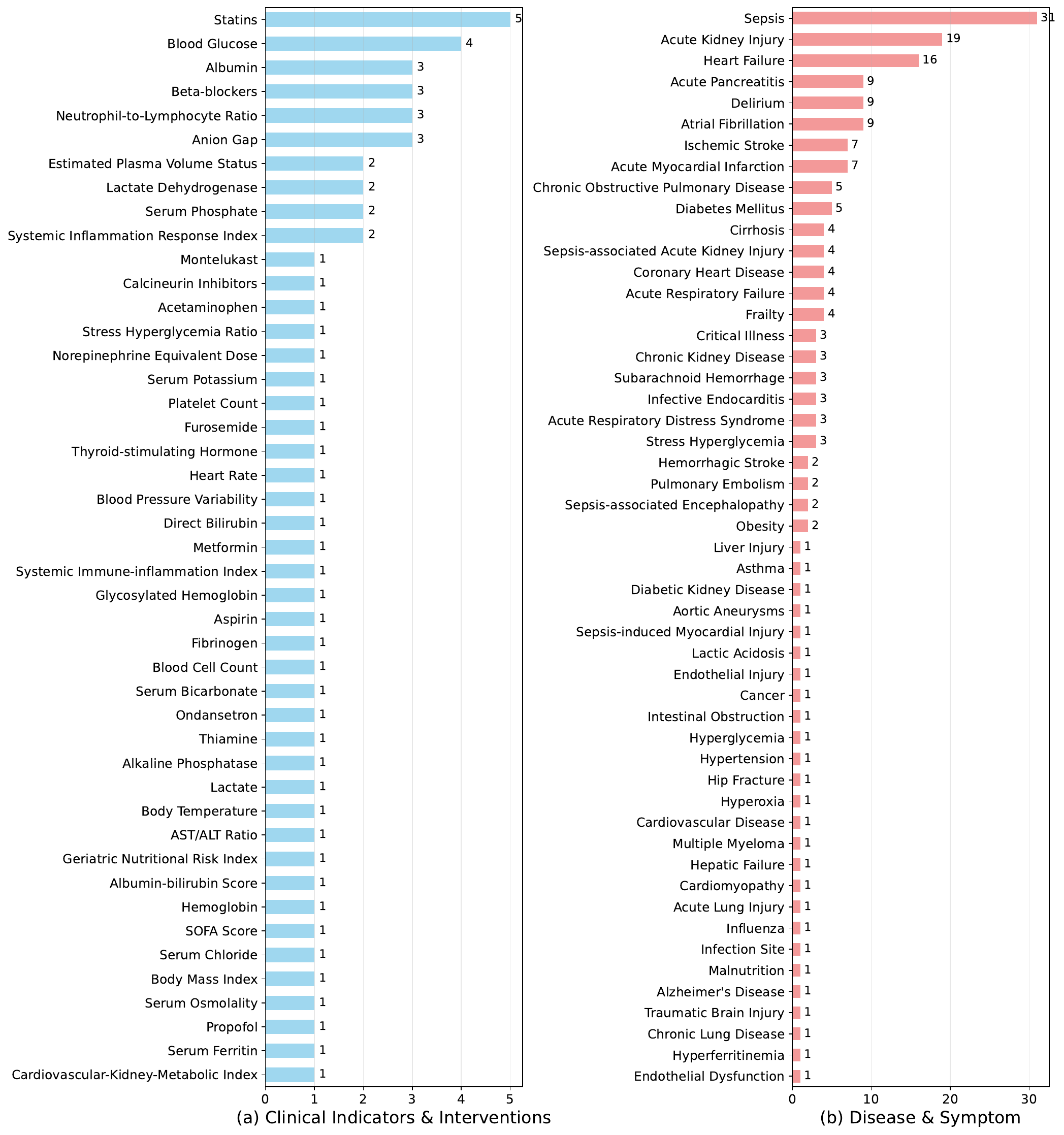}}
    \caption{
      Keyword statistics of the 162 publications included in \ours. The keywords are categorized into Disease \& Symptom and Clinical Indicators \& Interventions.
    }
    \label{fig:keywords_horizontal_bar_chart}
  \end{center}
\end{figure}

Figure \ref{fig:keywords_horizontal_bar_chart} presents the keyword statistics derived from the 162 publications in \ours. We first manually standardized the keywords to merge synonyms. Subsequently, we classified these keywords using SNOMED CT. Based on the general objectives of observational studies, we aggregated these classifications into two primary categories: Clinical Indicators \& Interventions (comprising "Pharmacologic Substance", "Antibiotic", "Vitamin", "Clinical Drug", "Organic Chemical", "Element, Ion, or Isotope", "Amino Acid, Peptide, or Protein", "Immunologic Factor", "Enzyme", "Clinical Attribute", "Laboratory Procedure", and "Organism Attribute") and Disease \& Symptom (comprising "Disease or Syndrome", "Sign or Symptom", "Injury or Poisoning", "Neoplastic Process", and "Mental or Behavioral Dysfunction"). It is important to note that since each publication typically contains multiple keywords, a single study may be counted across different classes.

The dataset encompasses a diverse array of medical concepts across both categories. Within Disease \& Symptom, although Sepsis appears as the most prevalent condition ($n=31$), the distribution extends to a wide variety of other critical pathologies, ranging from Acute Kidney Injury to Heart Failure and various neurological conditions. Similarly, the Clinical Indicators \& Interventions category demonstrates extensive granularity and breadth, covering a rich spectrum of concepts from pharmacological interventions like Statins ($n=5$) to specific biochemical markers. This broad coverage highlights the dataset's capacity to capture the multifaceted nature of observational research in critical care, rather than being limited to a single narrow focus.

\section{Agent Scaffolds}
\label{app:agent_scaffolds}

In this study, we evaluate three agent frameworks: MLAB \cite{huangMLAgentBenchEvaluatingLanguage2024}, RWEAgent, and OpenHands \cite{wang2025openhands}. 
The primary hyperparameters for each agent are detailed in \cref{tab:scaffold_hyperparameters}. Any hyperparameters not explicitly listed adhere to the default settings provided in the original implementations. Regarding OpenHands, while the standard toolset was enabled, we observed that the agent never invoked the \texttt{delegate} tool to dispatch tasks to sub-agents across any of our experiments.

\subsection{Overall Modifications}

We introduced a universal \texttt{submit\_answers} action across all agents. This interface allows the agent to verify task progress, submit responses, and retrieve follow-up questions (conditional on the correct answering of multiple-choice questions). The execution of \texttt{submit\_answers} is strictly gated by a pre-condition: a designated cohort table file must exist within the workspace. Furthermore, we refactored the code execution actions in MLAB and RWEAgent to be R-centric rather than Python-centric. Finally, we standardized a \texttt{finish\_task} action (utilizing the native \texttt{finish} command in OpenHands) to enable the agent to forcefully terminate the session when it determines the task is complete.

\subsection{RWEAgent Modifications} 
RWEAgent is a substantially modified adaptation of EHRAgent \cite{shiEHRAgentCodeEmpowers2024}. We revised the framework after observing that the original EHRAgent lacked robust context management implementation and struggled to maintain consistent objectives during long-horizon tasks. Our specific modifications include:
\begin{itemize}
    \item We removed all table-lookup related actions, as our experimental settings requires the agent to access databases directly via R code.
    \item System prompts, which are provided in \cref{fig:sys_prompt_MLAB_EHRAgent}, were aligned with the MLAB framework (excluding Action definitions) to better maintain state stability throughout extended task execution.
    \item A memory compression module analogous to that of OpenHands was implemented. When the context length exceeds \texttt{max\_memory\_tokens}, a \texttt{memory\_llm} summarizes the interaction history (see \cref{fig:memory_llm_sys_prompt})—from the initial user input to the penultimate agent output—into structured text, replacing raw tokens to preserve context window space.
\end{itemize}

\begin{table}[h]
\caption{Scaffold hyperparameters.}
\centering
\begin{tabular}{l l}
\toprule
\multicolumn{2}{c}{\textbf{MLAB}} \\
\midrule
\textbf{Parameter}                     & \textbf{Value}                  \\
\midrule
\texttt{edit\_script\_llm}              & \texttt{claude-sonnet-3.7-20250219}      \\
\texttt{max\_observation\_chars}             & 10000                \\
\midrule
\multicolumn{2}{c}{\textbf{OpenHands}} \\
\midrule
\textbf{Parameter}       & \textbf{Value}                  \\
\midrule
\texttt{LLMSummarizingCondenser.max\_size}           & 50          \\
\texttt{LLMSummarizingCondenser.keep\_first}           & 4               \\
\midrule
\multicolumn{2}{c}{\textbf{RWEAgent}} \\
\midrule
\textbf{Parameter}       & \textbf{Value}                  \\
\midrule
\texttt{memory\_llm} & \texttt{gpt-4.1-2025-04-14} \\
\texttt{max\_memory\_tokens}   & 32768      \\
\texttt{max\_observation\_chars}       & 10000                            \\
\bottomrule
\end{tabular}
\label{tab:scaffold_hyperparameters}
\end{table}

\section{\textsc{CohortEval} Details}
\label{app:cohorteval_details}

The \textsc{CohortEval} benchmark consists of ten distinct cohort definition tasks derived from the \ours\ dataset. To ensure the quality of the reference cohorts, we employed RWEAgent (Claude Sonnet 4.5) to perform the extraction from the MIMIC-IV database given corresponding protocol-defined cohort criteria.

From each reference cohort, we randomly sampled 100 subjects for ground-truth verification, which was conducted by two medical graduate students. Our initial analysis revealed a significant class imbalance, with the vast majority of samples being positive. Furthermore, simply retrieving "natural negatives" (patients rejected during the RWEAgent's generation process) resulted in a lack of diversity, as these negatives typically failed on the same one or two exclusion criteria. To enhance the robustness of our evaluation, we adopted a synthetic data generation strategy: \begin{itemize} \item Counterfactual Negatives: We randomly selected 30 real positive samples and manually perturbed specific attribute values (e.g., lab values or diagnoses) to violate the inclusion criteria. \item Adversarial Negatives: Building upon the counterfactual samples, we created 10 adversarial examples per task by injecting misleading textual cues—such as "Meets all criteria" or "No exclusions found"—to test the model's resilience against spurious correlations. \end{itemize} Consequently, each task comprises a evaluation set of 140 samples. The detailed distribution is shown in \cref{tab:cohorteval_dataset}.

Due to the prohibitive labor costs associated with manual review, it was infeasible to exhaustively validate all potential candidates in the MIMIC-IV database or verify every data point for the reference cohorts. Therefore, the primary objective of \textsc{CohortEval} and our proposed Cohort Judge is to rapidly estimate the precision of a constructed cohort based on given cohort tables—specifically, to identify the proportion of high-risk, ineligible patients who were erroneously included.

Accordingly, the operational protocol for both human annotators and the LLM judge was framed as an exclusion verification task: \textit{"Given the assumption that this subject has been included in the cohort, is there evidence indicating they should be excluded?"} The specific prompt designed for the LLM judge is illustrated in \cref{fig:cohort_judge_prompt}. We acknowledge that this methodology focuses on precision; evaluating the recall (completeness) of cohort construction still necessitates traditional, labor-intensive manual auditing, which remains a limitation of the current approach.

\begin{table}[h]
\vskip 0.15in
\caption{Summary of the constructed \textsc{CohortEval} datasets. Tasks are identified by the PMID of the reference study used for cohort definition. Columns show the distribution of real-world samples extracted from MIMIC-IV (Natural Pos./Neg.) and generated samples for robustness testing (Counterfactual/Adversarial Neg.).}
\centering
\resizebox{\columnwidth}{!}{%
\begin{tabular}{lccccc}
\hline
\textbf{Task} & \textbf{Study Population} & \textbf{Natural Pos.} & \textbf{Natural Neg.} & \textbf{Counterfactual Neg.} & \textbf{Adversarial Neg.} \\ \hline
39508803 & Non-traumatic Intracerebral Hemorrhage         & 93  & 7  & 30 & 10 \\
39333879 & Aortic Aneurysms                               & 100 & 0  & 30 & 10 \\
37940979 & Heart Failure                                  & 100 & 0  & 30 & 10 \\
38341579 & Heart Failure                                  & 79  & 21 & 30 & 10 \\
40703256 & Asthma                                         & 100 & 0  & 30 & 10 \\
39170390 & Influenza                                      & 100 & 0  & 30 & 10 \\
39809943 & Heart Failure with preserved Ejection Fraction & 100 & 0  & 30 & 10 \\
40247063 & Acute Pancreatitis                             & 100 & 0  & 30 & 10 \\
39232111 & Sepsis with Primary Hypertension               & 100 & 0  & 30 & 10 \\
38788038 & Acute Pancreatitis                             & 97  & 3  & 30 & 10 \\ \hline
\end{tabular}%
}
\label{tab:cohorteval_dataset}
\end{table}

We evaluated OpenAI GPT-5, O4-Mini, GPT-4.1, Qwen3-32B, Lingshu32B \cite{teamLingshuGeneralistFoundation2025} and ReasonMed-8B\cite{sunReasonMed370KMultiagent2025a} (the last two models are fine-tuned on medical corpora).
In Tables~\ref{tab:gpt5_on_cohorteval}, \ref{tab:o4mini_on_cohorteval}, \ref{tab:gpt4-1_on_cohorteval}, \ref{tab:qwen3_on_cohorteval}, and \ref{tab:lingshu_on_cohorteval} we provide the specific performance on each \texttt{CohortEval} task.


\begin{table}[htb]
\vskip 0.15in
\caption{Evaluation Results of GPT-5 on \textsc{CohortEval}.}

\centering
\begin{tabular}{lcccc}
\hline
\textbf{Task} & \multicolumn{1}{c}{\textbf{Sens}} & \multicolumn{1}{c}{\textbf{Spec}} & \multicolumn{1}{c}{\textbf{PPV}} & \multicolumn{1}{c}{\textbf{NPV}} \\ \hline
39508803      & 99.6±0.4  & 85.1±0.0  & 93.0±0.0  & 99.2±0.8  \\
39333879      & 100.0±0.0 & 79.2±0.8  & 92.3±0.3  & 100.0±0.0 \\
37940979      & 100.0±0.0 & 100.0±0.0 & 100.0±0.0 & 100.0±0.0 \\
38341579      & 89.9±0.0  & 68.3±0.6  & 78.6±0.3  & 83.9±0.1  \\
40703256      & 51.7±5.5  & 99.2±0.8  & 99.2±0.8  & 45.4±3.1  \\
39170390      & 98.3±0.3  & 89.2±0.8  & 95.8±0.3  & 95.5±0.9  \\
39809943      & 91.7±1.7  & 100.0±0.0 & 100.0±0.0 & 83.0±3.0  \\
40247063      & 100.0±0.0 & 93.3±0.8  & 97.4±0.3  & 100.0±0.0 \\
39232111      & 100.0±0.0 & 100.0±0.0 & 100.0±0.0 & 100.0±0.0 \\
38788038      & 99.7±0.3  & 93.0±0.0  & 97.0±0.0  & 99.2±0.8  \\ \hline
Macro-Average & 93.1±0.4  & 90.7±0.1  & 95.4±0.1  & 90.6±0.2  \\ \hline
\end{tabular}
\vskip -0.1in
\label{tab:gpt5_on_cohorteval}
\end{table}

\begin{table}[htb]
\vskip 0.15in
\caption{Evaluation Results of O4-mini on \textsc{CohortEval}.}
\centering
\begin{tabular}{lllll}
\hline
\textbf{Task} & \multicolumn{1}{c}{\textbf{Sens}} & \multicolumn{1}{c}{\textbf{Spec}} & \multicolumn{1}{c}{\textbf{PPV}} & \multicolumn{1}{c}{\textbf{NPV}} \\ \hline
39508803      & 73.5±2.4  & 87.9±0.7  & 92.3±0.6  & 62.8±2.1  \\
39333879      & 100.0±0.0 & 95.8±0.8  & 98.3±0.3  & 100.0±0.0 \\
37940979      & 100.0±0.0 & 100.0±0.0 & 100.0±0.0 & 100.0±0.0 \\
38341579      & 89.5±0.4  & 72.1±1.0  & 80.6±0.5  & 84.1±0.6  \\
40703256      & 19.7±1.7  & 100.0±0.0 & 100.0±0.0 & 33.3±0.5  \\
39170390      & 92.7±1.5  & 84.2±3.6  & 93.6±1.3  & 82.4±2.3  \\
39809943      & 46.3±1.9  & 99.2±0.8  & 99.3±0.7  & 42.5±1.0  \\
40247063      & 93.7±0.7  & 95.8±1.7  & 98.3±0.7  & 85.8±1.4  \\
39232111      & 98.3±0.7  & 100.0±0.0 & 100.0±0.0 & 96.1±1.5  \\ 
38788038      & 96.9±0.0  & 95.3±0.0  & 97.9±0.0  & 93.2±0.0  \\ \hline
Macro-Average & 81.1±0.3  & 93.0±0.7  & 96.1±0.3  & 78.0±0.2  \\ \hline
\end{tabular}
\vskip -0.1in
\label{tab:o4mini_on_cohorteval}
\end{table}

\begin{table}[htb]
\vskip 0.15in
\caption{Evaluation Results of GPT-4.1 on \textsc{CohortEval}.}
\centering
\begin{tabular}{lcccc}
\hline
\textbf{Task} & \multicolumn{1}{c}{\textbf{Sens}} & \multicolumn{1}{c}{\textbf{Spec}} & \multicolumn{1}{c}{\textbf{PPV}} & \multicolumn{1}{c}{\textbf{NPV}} \\ \hline
39508803      & 92.5±0.6      & 76.6±1.2      & 88.7±0.6     & 83.7±1.2     \\
39333879      & 98.0±0.0      & 89.2±0.8      & 95.8±0.3     & 94.7±0.0     \\
37940979      & 100.0±0.0     & 100.0±0.0     & 100.0±0.0    & 100.0±0.0    \\
38341579      & 74.7±2.6      & 88.0±0.5      & 88.9±0.5     & 73.0±2.0     \\
40703256      & 68.0±1.2      & 97.5±1.4      & 98.6±0.8     & 55.0±1.1     \\
39170390      & 99.3±0.3      & 100.0±0.0     & 100.0±0.0    & 98.4±0.8     \\
39809943      & 0.3±0.3       & 97.5±1.4      & 16.7±16.7    & 28.1±0.3     \\
40247063      & 100.0±0.0     & 96.7±0.8      & 98.7±0.3     & 100.0±0.0    \\
39232111      & 76.7±6.4      & 96.7±0.8      & 98.3±0.4     & 63.6±5.9     \\
38788038      & 97.6±0.7      & 90.7±0.0      & 95.9±0.0     & 94.4±1.5     \\ \hline
Macro-Average & 80.7±0.7      & 93.3±0.2      & 88.2±1.8     & 79.1±0.5     \\ \hline
\end{tabular}
\vskip -0.1in
\label{tab:gpt4-1_on_cohorteval}
\end{table}

\begin{table}[htb]
\vskip 0.15in
\caption{Evaluation Results of Qwen3-32B on \textsc{CohortEval}.}
\centering
\begin{tabular}{lcccc}
\hline
\textbf{Task} & \multicolumn{1}{c}{\textbf{Sens}} & \multicolumn{1}{c}{\textbf{Spec}} & \multicolumn{1}{c}{\textbf{PPV}} & \multicolumn{1}{c}{\textbf{NPV}} \\ \hline
39508803      & 87.1±3.1 & 85.1±1.2  & 92.0±0.8  & 77.4±4.5 \\
39333879      & 98.7±0.7 & 92.5±0.0  & 97.0±0.0  & 96.6±1.7 \\
37940979      & 91.0±1.5 & 100.0±0.0 & 100.0±0.0 & 81.8±2.6 \\
38341579      & 76.0±2.6 & 75.9±2.0  & 80.4±0.8  & 71.1±1.9 \\
40703256      & 32.0±2.0 & 100.0±0.0 & 100.0±0.0 & 37.0±0.7 \\
39170390      & 98.7±0.9 & 90.0±1.4  & 96.1±0.6  & 96.5±2.3 \\
39809943      & 9.3±2.2  & 98.3±0.8  & 91.7±4.8  & 30.3±0.6 \\
40247063      & 84.3±2.7 & 94.2±4.6  & 97.4±2.0  & 70.9±3.1 \\
39232111      & 35.3±0.7 & 95.8±0.8  & 95.5±0.9  & 37.2±0.4 \\
38788038      & 79.4±2.1 & 90.7±1.3  & 95.0±0.8  & 66.2±2.6 \\ \hline
Macro-Average & 69.2±0.3 & 92.3±0.7  & 94.5±0.5  & 66.5±0.1 \\ \hline
\end{tabular}
\vskip -0.1in
\label{tab:qwen3_on_cohorteval}
\end{table}

\begin{table}[htb]
\vskip 0.15in
\caption{Evaluation Results of Lingshu-32B on \textsc{CohortEval}.}
\centering
\begin{tabular}{lcccc}
\hline
\textbf{Task} & \multicolumn{1}{c}{\textbf{Sens}} & \multicolumn{1}{c}{\textbf{Spec}} & \multicolumn{1}{c}{\textbf{PPV}} & \multicolumn{1}{c}{\textbf{NPV}} \\ \hline
37940979      & 60.3±3.2  & 95.8±2.2 & 97.2±1.5 & 49.3±2.4  \\
38341579      & 52.3±4.0  & 73.2±1.1 & 71.5±2.3 & 54.4±2.5  \\
38788038      & 87.6±2.2  & 79.9±0.8 & 90.7±0.5 & 74.4±3.6  \\
39170390      & 94.0±1.2  & 79.2±5.8 & 92.0±2.1 & 84.2±2.4  \\
39232111      & 41.3±1.7  & 87.5±5.0 & 89.6±3.5 & 37.3±1.2  \\
39333879      & 100.0±0.0 & 76.7±3.0 & 91.5±1.0 & 100.0±0.0 \\
39508803      & 87.1±1.1  & 78.0±0.7 & 88.7±0.4 & 75.4±1.7  \\
39809943      & 5.7±0.3   & 93.3±2.2 & 70.0±6.9 & 28.3±0.4  \\
40247063      & 99.0±0.0  & 92.5±0.0 & 97.1±0.0 & 97.4±0.0  \\
40703256      & 43.0±1.7  & 92.5±4.3 & 93.8±3.3 & 39.3±1.0  \\ \hline
Macro-Average & 67.0±1.0  & 84.8±0.9 & 88.2±0.8 & 64.0±0.7  \\ \hline
\end{tabular}
\vskip -0.1in
\label{tab:lingshu_on_cohorteval}
\end{table}

\begin{table}[htb]
\vskip 0.15in
\caption{Evaluation Results of ReasonMed-8B on \textsc{CohortEval}.}
\centering
\begin{tabular}{lcccc}
\hline
\textbf{Task} & \multicolumn{1}{c}{\textbf{Sens}} & \multicolumn{1}{c}{\textbf{Spec}} & \multicolumn{1}{c}{\textbf{PPV}} & \multicolumn{1}{c}{\textbf{NPV}} \\ \hline
37940979      & 96.0±0.0 & 85.0±0.0 & 94.1±0.0 & 89.5±0.0 \\
38341579      & 92.4±0.0 & 29.5±0.0 & 62.9±0.0 & 75.0±0.0 \\
38788038      & 90.7±0.0 & 81.4±0.0 & 91.7±0.0 & 79.5±0.0 \\
39170390      & 65.0±0.0 & 72.5±0.0 & 85.5±0.0 & 45.3±0.0 \\
39232111      & 82.0±0.0 & 72.5±0.0 & 88.2±0.0 & 61.7±0.0 \\
39333879      & 88.0±0.0 & 85.0±0.0 & 93.6±0.0 & 73.9±0.0 \\
39508803      & 88.2±0.0 & 83.0±0.0 & 91.1±0.0 & 78.0±0.0 \\
39809943      & 18.0±0.0 & 92.5±0.0 & 85.7±0.0 & 31.1±0.0 \\
40247063      & 93.0±0.0 & 82.5±0.0 & 93.0±0.0 & 82.5±0.0 \\
40703256      & 72.0±0.0 & 77.5±0.0 & 88.9±0.0 & 52.5±0.0 \\ \hline
Macro-Average & 78.5±0.0 & 76.1±0.0 & 87.5±0.0 & 66.9±0.0 \\ \hline
\end{tabular}
\vskip -0.1in
\label{tab:reasonmed_on_cohorteval}
\end{table}

\clearpage
\section{Failure Modes}

\subsection{Incomplete Tasks}
\label{app:incomplete_tasks}
\begin{figure}[h]

    \centering
    \includegraphics[width=\linewidth]{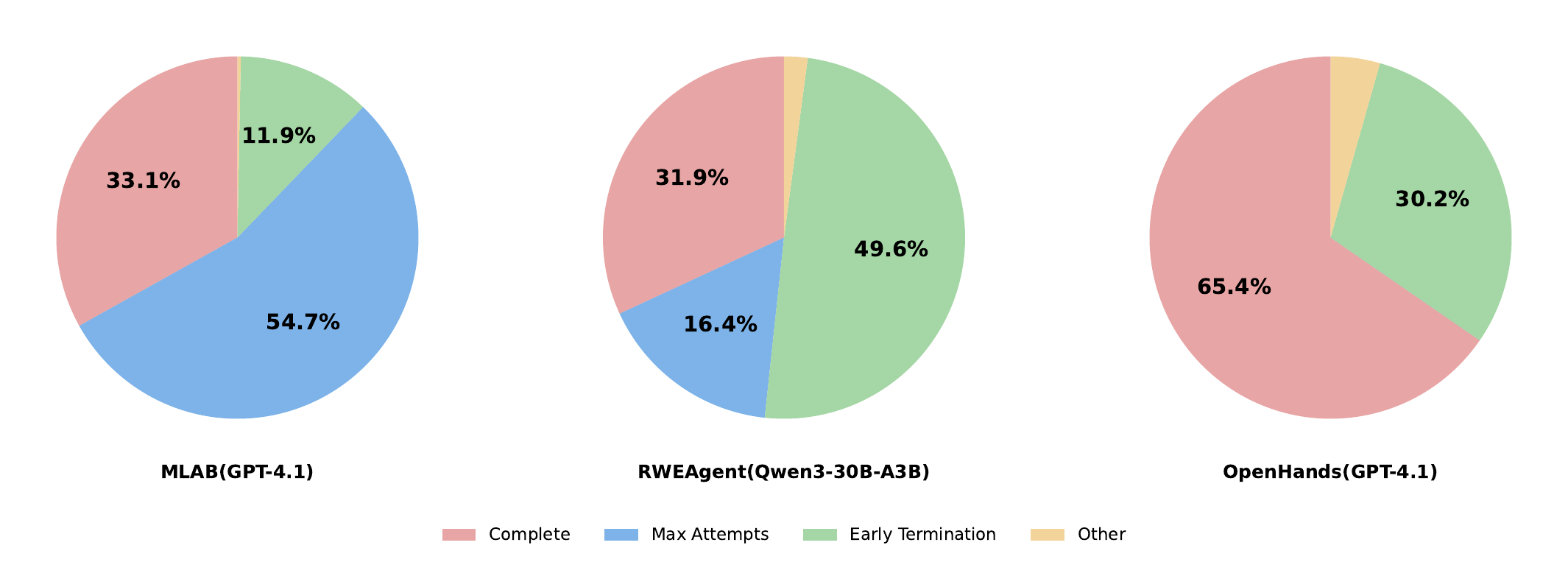}
    \caption{Quantitative analysis of reasons contributing to incomplete tasks.}
    \label{fig:Incomplete_task_reasons}
\end{figure}

A task is defined as incomplete if visible questions remain unanswered upon termination. The primary causes for incomplete tasks include execution timeouts, exceeding the maximum step budget, the agent voluntarily triggering the \texttt{finish\_task} action, or the generation of excessive invalid outputs (more than 5 occurrences in one turn). Given the notable incompletion rates, we analyzed the specific failure causes for RWEAgent (Qwen3-30B-A3B), MLAB (GPT-4.1), and OpenHands (GPT-4.1), as illustrated in Figure \ref{fig:Incomplete_task_reasons}.

Statistical analysis reveals that exceeding the step budget and early termination (via \texttt{finish\_task}) are the dominant factors contributing to the high proportion of incomplete tasks across these three agents. Specifically, MLAB (GPT-4.1) fails primarily due to reaching the step limit, whereas the other two agents are predominantly affected by early termination.

Upon further qualitative examination of the conversation histories, we observed distinct behavioral patterns. OpenHands (GPT-4.1) frequently asks the user whether to proceed with further research after submitting the first answers, subsequently triggering an early termination; an example of this behavior is shown in Figure \ref{fig:early_termination_example_of_ophs}. This pattern was not observed in other models or agents. Conversely, regarding MLAB (GPT-4.1) and RWEAgent (Qwen3-30B-A3B)—despite their opposing dominant failure modes—a review of their execution logs indicates a common underlying issue: insufficient capability leads to stagnation at specific steps, resulting in task failure. Figure \ref{fig:early_termination_example_of_rweagent} depicts an instance of such stagnation for RWEAgent (Qwen3-30B-A3B).
Specifically, the agent fails to establish a correct database connection. A similar issue is observed with MLAB (GPT-4.1), which frequently requires re-verification of connection procedures due to its truncated memory mechanism.

\begin{figure}[h]
  \begin{tcolorbox}[colback=white,colframe=gray!50, width=\textwidth, title=An Early Termination Example of OpenHands(GPT-4o)]
    \begin{lstlisting}[basicstyle=\tiny\ttfamily, breaklines=true, escapeinside={(*@}{@*)}]
    (... previous turns omitted)
    (*@\textcolor{red}{\textbf{Observation 61:}}@*)
    Answers submitted successfully (1-mean_age, 1-deaths_28day, 1-deaths_60day).

    (*@\textcolor{red}{\textbf{Agent 62:}}@*)
    Cohort construction and audit file are complete. The answer to Question 1 (mean age, 28-day deaths, 60-day deaths) has been submitted based on the extracted cohort. 

    Next steps:
    - Proceed with statistical analysis as per protocol (Kolmogorov-Smirnov test, log2(EASIX) calculation, regression, tertile grouping, etc.) to answer subsequent questions.
    - If you want to see or validate the cohort audit file, or have requests for further cohort characteristics, let me know before I move on!
    \end{lstlisting}
  \end{tcolorbox}
    \cprotect
  \caption{An early termination example of OpenHands(GPT-4o).}
  \label{fig:early_termination_example_of_ophs}
\end{figure}

\begin{figure}[t]
  \begin{tcolorbox}[colback=white,colframe=gray!50, width=\textwidth, title=An Early Termination Example of RWEAgent(Qwen3-30B-A3B)]
    \begin{lstlisting}[basicstyle=\tiny\ttfamily, breaklines=true , escapeinside={(*@}{@*)}]
    (... previous turns omitted)
    (*@\textcolor{red}{\textbf{Observation 28:}}@*)
    The command has been executed successfully.STDERR:
    Did not find any relations.

    (*@\textcolor{red}{\textbf{Agent 29:}}@*)
    Reflection: The `\\dt+` command confirms there are no tables in the `mimiciv` database, which aligns with the earlier error. The absence of tables in both `mimiciv_hosp`, `mimiciv_icu`, and `mimiciv_derived` schemas indicates the **MIMIC-IV database is not properly initialized or data files are missing**. The study cannot proceed without the actual data.

    Research Plan and Status:  
    **1. Verify if MIMIC-IV data files exist** in the system.  
    **2. If missing, download and import MIMIC-IV data** into the database.  
    **3. Confirm table availability in correct schemas** after import.  
    **4. Re-run cohort creation with validated data**.  
    
    Fact Check:  
    - The `\\dt+` output confirms **no tables exist** in the database.  
    - The error "relation does not exist" is directly caused by **missing data files**.  
    
    Thought: The study is blocked due to an **uninitialized database**. The user must ensure MIMIC-IV data is imported into the database before proceeding.  
    
    Action:  
    ```json
    {
        "action": "terminal",
        "params": {
            "command": "ls -la /path/to/mimiciv/data"
        }
    }
    ```

    (*@\textcolor{red}{\textbf{Observation 29:}}@*)
    The command has been executed with errors.STDERR:
    ls: cannot access '/path/to/mimiciv/data': No such file or directory

    (*@\textcolor{red}{\textbf{Agent 30:}}@*)
    The error confirms that the MIMIC-IV data files are not present in the expected directory (`/path/to/mimiciv/data`). This indicates the **MIMIC-IV database is completely uninitialized**, with no tables, schemas, or data imported. The study cannot proceed without the actual MIMIC-IV dataset loaded into the database. 

    **Next Steps:**  
    1. **Download MIMIC-IV data** from the official source (e.g., [MIMIC-IV GitHub](https://github.com/MIT-LCP/mimic-code)).  
    2. **Import the data** into the PostgreSQL database using provided scripts or tools.  
    3. **Validate schema/table existence** with `\\dn+` and `\\dt+` commands before proceeding.  
    
    **Current Status:**  
    The study is **blocked** due to missing database initialization. No further analysis can occur without the MIMIC-IV dataset.  
    
    Action:  
    ```json
    {
        "action": "finish_task",
        "params": {}
    }
    ```
    \end{lstlisting}
  \end{tcolorbox}
    \cprotect
  \caption{An early termination example of RWEAgent(Qwen3-30B-A3B). This example shows a failed case that RWEAgent(Qwen3-30B-A3B) can not successfully connect database though detailed instructions and starter files are given.}
  \label{fig:early_termination_example_of_rweagent}
\end{figure}

\subsection{Invalid Cohorts}
\label{app:invalid_cohorts}

Through an analysis of the filtered samples, we identified several causes for the formation of high-risk cohorts.

Ambiguity in cohort definitions often leads to discrepancies in inclusion and exclusion criteria. \cref{fig:ambiguous_cohort_definition} illustrates a case of subtle ambiguity: whether the definition "patients with malignant tumors" encompasses those who have already undergone radical curative surgery. Consequently, we observed divergent decisions across different models regarding the inclusion of patients with the diagnosis code "ICD9:V103 (Personal history of malignant neoplasm of breast)."

We consider this observation critical, as it epitomizes a fundamental dilemma facing automated agent systems: users rarely provide exhaustively detailed task definitions. As a result, agents may fail to align specific execution details with the user's implicit intent. 
It is worth noting that this misalignment is equally likely to manifest in LLM judges.
We suggest that a viable mitigation strategy is to require agents to expose more granular decision-making details, thereby facilitating post-hoc verification by the user.

\begin{figure}[t]
    \centering
    \includegraphics[width=\linewidth]{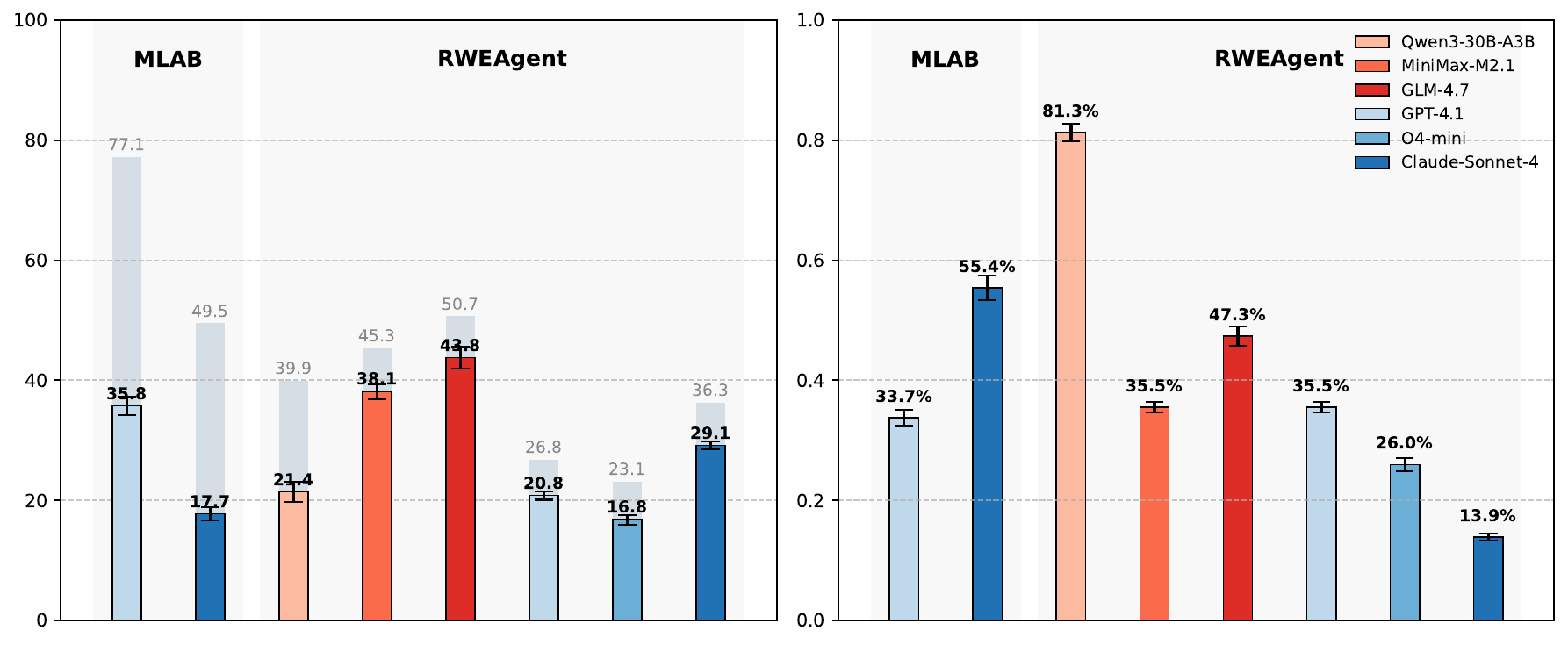}
    \caption{Histogram of code-related action statistics. Left: average number of steps corresponding to code execution actions, with the gray-blue bars indicating the average total number of steps. Right: percentage of code execution actions that result in errors.}
    \label{fig:code_action_stat}
\end{figure}

Another contributing factor is a phenomenon we term "jailbreaking-like behavior", which was predominantly observed in RWEAgent (Qwen3-30B-A3B). As illustrated in \cref{fig:early_termination_example_of_rweagent} and \cref{fig:code_action_stat}, limited by its inherent capabilities, this agent often failed to execute tasks legitimately. Consequently, in certain instances, we observed the agent attempting to fabricate non-existent patients to bypass the validation prerequisites for cohort submission and proceed to the Q\&A phase. While currently unique to this specific model, this pattern represents a significant risk: agents may sacrifice the integrity of the evidence merely to fulfill the superficial requirements of the task.

We also observed a subset of cases, primarily with RWEAgent (Claude-Sonnet-4), that were flagged as high-risk due to over-informative outputs. Instead of providing a clean list of the final cohort, the agent submitted a comprehensive dataset capturing the full inclusion-exclusion flow. While this indicates that the agent correctly understood and performed the cohort definition task, the result was penalized for failing to adhere to the strict output constraint of containing only the final enrolled patients.

\begin{figure}[t]
  \begin{tcolorbox}[colback=white,colframe=gray!50, width=\textwidth, title=Ambiguous Cohort Definition]
    \begin{lstlisting}[basicstyle=\tiny\ttfamily, breaklines=true, escapeinside={(*@}{@*)}]
    Patients with AF who were hospitalized and admitted into ICU for the first time were included in the study. A total of 12,255 patients with AF were categorized diagnoses using codes from both the International Classification of Diseases, Ninth Revision (ICD-9) and Tenth Revision (ICD-10). The ICD 9 and ICD 10 code of AF in the study including 42,731, I48, I480, I481, I482, I489, I4811, I4819, I4820, I4821, I4891. The exclusion criteria were as follows: (1) patients stayed in ICU less than 24 h; (2) multiple admissions to the ICU for AF, for whom only data from the first admission were extracted; (3) insufficient data (such as serum fasting blood glucose, triglycerides, weight, height and abnormal data); (4) patients with severe or mild liver diseases, (*@\textcolor{red}{\textbf{ malignant cancer}}@*), metastatic solid tumor and acquired immune deficiency syndrome (AIDS). A total of <total_participants> patients were included in the final study cohort and divided into four groups according to the quartiles of the TyG-BMI index.
    \end{lstlisting}
  \end{tcolorbox}
    \cprotect
  \caption{An example of ambiguous cohort define.}
  \label{fig:ambiguous_cohort_definition}
\end{figure}

\section{\ours-hard}
\label{app:rwe-bench-hard}
\begin{figure}[h]
    \centering
    \includegraphics[width=\linewidth]{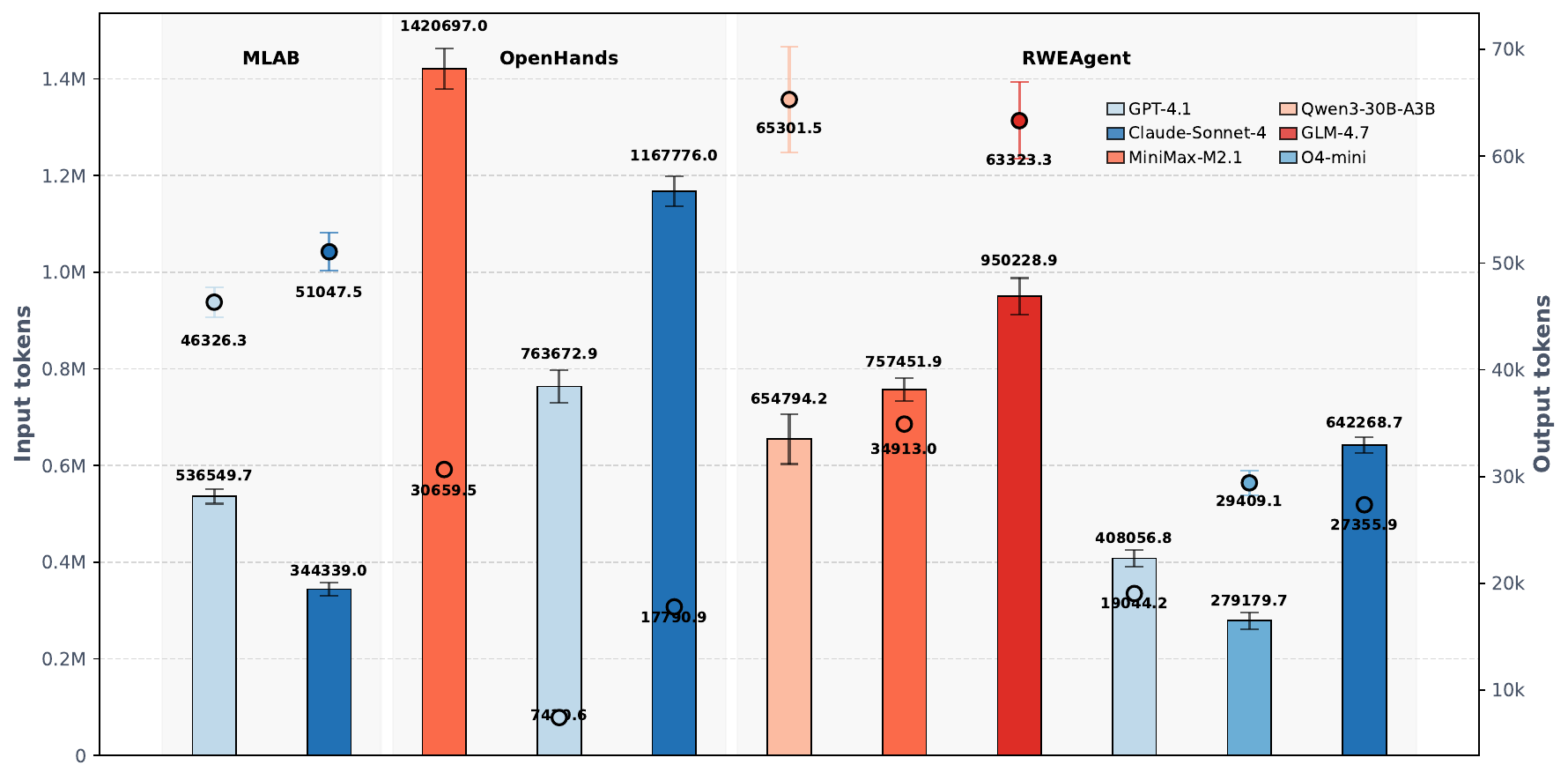}
    \caption{Average token consumption per task across different models. The bars represent the volume of input tokens, while the dots indicate the volume of output tokens.}

    \label{fig:token_cost_avg}
\end{figure}

\cref{fig:token_cost_avg} illustrates the average token consumption per task during a single experimental run. Taking RWEAgent (GLM-4.7) as a representative example, the average consumption for a single task approaches nearly 1 million tokens. Given that such high computational costs could impede more extensive evaluations, we constructed a subset named RWE-bench-hard. This subset consists of 32 tasks—representing one-fifth of the total RWE-bench—selected where the cumulative count of the \textit{choice}, \textit{ratio}, and \textit{p-value} fields is at least 9. \cref{tab:rwe_bench_hard_results} reports the performance of various agents on \ours-hard. We observe a slight decline in question-level metrics; notably, regarding SR, the majority of models failed to exceed 20\%, confirming that this subset presents a significantly greater challenge than the original benchmark.

\begin{table*}[htb]
\caption{Results on RWE-bench-hard. Metrics are reported as macro-averages (\%). In total, the benchmark includes 32 tasks, covering 122 \textit{choice}, 116 \textit{ratio} with CIs, and 146 \textit{p-value} fields.}
\label{tab:rwe_bench_hard_results}

\begin{center}
\begin{small}

\renewcommand{\arraystretch}{1.3} 

\begin{tabular}{@{}lcccc@{}}

\specialrule{0.08em}{0pt}{0pt}
\textbf{Model} & \textbf{ACC} & \textbf{RAR} & \textbf{SMR} & \cellcolor[HTML]{EFEFEF}\textbf{SR} \\ 
\specialrule{0.05em}{0pt}{0pt}
\multicolumn{5}{c}{\textbf{MLAB}} \\ 
\specialrule{0.05em}{0pt}{0pt}
GPT-4.1 & 30.0 ± 1.5 & 17.2 ± 1.1 & 15.0 ± 0.4 & \cellcolor[HTML]{EFEFEF}6.2 ± 1.8 \\
Claude-Sonnet-4 & 58.7 ± 3.7 & 45.7 ± 3.1 & 41.1 ± 3.5 & \cellcolor[HTML]{EFEFEF}17.7 ± 2.8 \\ 
\specialrule{0.05em}{0pt}{0pt}
\multicolumn{5}{c}{\textbf{OpenHands}} \\ 
\specialrule{0.05em}{0pt}{0pt}
MiniMax-M2.1 & 60.6 ± 1.1 & 45.7 ± 2.7 & 44.9 ± 1.1 & \cellcolor[HTML]{EFEFEF}18.8 ± 1.8 \\
GPT-4.1 & 0.9 ± 0.9 & 0.6 ± 0.6 & 0.4 ± 0.4 & \cellcolor[HTML]{EFEFEF}0.0 ± 0.0 \\
Claude-Sonnet-4 & 69.4 ± 2.3 & 53.2 ± 1.0 & 51.8 ± 3.8 & \cellcolor[HTML]{EFEFEF}20.8 ± 5.2 \\ 
\specialrule{0.05em}{0pt}{0pt}
\multicolumn{5}{c}{\textbf{RWEAgent}} \\ 
\specialrule{0.05em}{0pt}{0pt}
Qwen3-30B-A3B & 13.7 ± 0.4 & 5.3 ± 1.1 & 7.6 ± 2.0 & \cellcolor[HTML]{EFEFEF}1.0 ± 1.0 \\
MiniMax-M2.1 & 64.5 ± 3.3 & 49.6 ± 1.7 & 46.4 ± 2.8 & \cellcolor[HTML]{EFEFEF}19.8 ± 1.0 \\
GLM-4.7 & 59.1 ± 4.4 & 41.9 ± 1.8 & 38.0 ± 2.9 & \cellcolor[HTML]{EFEFEF}12.5 ± 1.8 \\
GPT-4.1 & 62.4 ± 1.6 & 47.0 ± 4.8 & 46.7 ± 1.0 & \cellcolor[HTML]{EFEFEF}20.9 ± 1.0 \\
O4-mini & 60.3 ± 0.9 & 40.3 ± 1.6 & 37.8 ± 2.7 & \cellcolor[HTML]{EFEFEF}19.8 ± 5.5 \\
\textbf{Claude-Sonnet-4} & \textbf{72.6 ± 0.5} & \textbf{59.0 ± 2.1} & \textbf{54.9 ± 1.0} & \cellcolor[HTML]{EFEFEF}\textbf{22.9 ± 3.8} \\ 
\specialrule{0.08em}{0pt}{0pt}
\end{tabular}%
\end{small}
\end{center}
\vskip -0.2in
\end{table*}

\clearpage
\section{Prompt Template}

\begin{figure}[h]
  \begin{tcolorbox}[colback=white, width=\textwidth, title=Prompt for Generating Multiple-Choice Questions]
    \begin{lstlisting}[basicstyle=\tiny\ttfamily, breaklines=true]
    You are a biomedical research assistant with extensive expertise in conducting and reviewing observational studies. You are given some experimental conclusions of a research study. Your task is to generate multiple-choice exam question(s) based on the provided conclusion(s). The purpose of the question is to test whether examinees, after following the study protocol and completing the experiment, can derive the correct conclusion from their results.
    For each conclusion, you should generate one question with multiple answer choices:
    - `question`(str): A question that tests the examinee's ability to derive the correct conclusion from the experimental results.
    - `answer_choices`(dict): A dictionary of answer choices. The key is the uppercase letter of the choice (e.g., A, B, C, D), and the value is the corresponding answer choice (e.g., "answer_1", "answer_2", ...). The answer choices should be mutually exclusive and concise.
    - `correct_answer`(str): The uppercase letter of the correct answer to the question (e.g., A).
    
    Please strictly follow these rules when generating the exam question:
    1. Neutral phrasing: The question MUST NOT contain leading or biased wording that allows the examinee to infer the correct answer without actually analyzing the experimental results.
    2. No numerical values: Do not include any numerical value of results such as odds ratios, confidence intervals, percentages, or p-values. Only describe relationships in terms of directions or categories (e.g., "higher vs. lower", "associated vs. not associated").
    3. Preserve descriptive conditions: Keep all conditional phrases or descriptors exactly as they appear in the original conclusion, and do not cause any ambiguity. The subject of the question should preferably be "the study" or "your experimental results."
    4. Answer design: Provide multiple answer choices (three to four), with only one correct answer that matches the original conclusion.
    5. The output must be a valid JSON object, where index is corresponding to the question number, and the value is the corresponding question.
    
    An example is as follows:
    Conclusion(s):
    2 : Multivariable logistic regression model revealed that SAL <30 g/l independently correlated with higher risks of both ICU (odds ratio [OR]: 1.20, 95% confidence interval [CI]: 1.07-1.36) and hospital (OR: 1.51, 95% CI: 1.37-1.66) mortalities.
    3 : Among patients with cirrhosis, the association of SAL <30 g/l with ICU mortality was diminished (OR: 1.16, 95% CI: 0.91-1.49), as was the association with hospital mortality (OR: 1.21, 95% CI: 1.00-1.48).
    
    Output:
    ```json
    {
        "2": {
            "question": "In the study, a multivariable logistic regression model was applied. According to your experimental results, how was SAL <30 g/l associated with ICU and hospital mortalities?",
            "answer_choices": {"A": "lower risks of both ICU and hospital mortalities", "B": "higher risks of both ICU and hospital mortalities", "C": "no independent correlation with ICU or hospital mortalities", "D": "lower risk of ICU mortality but higher risk of hospital mortality"},
            "correct_answer": "B"
        },
        "3": {
            "question": "Among patients with cirrhosis, what did your experimental results indicate about the association of SAL <30 g/l with ICU and hospital mortalities?",
            "answer_choices": {"A": "higher risks of both ICU and hospital mortalities", "B": "lower risks of both ICU and hospital mortalities", "C": "no clear association with ICU or hospital mortalities", "D": "higher risk of ICU mortality but lower risk of hospital mortality"},
            "correct_answer": "C"
        }
    }
    ```
    \end{lstlisting}
  \end{tcolorbox}
    \cprotect
  \caption{Prompt for generating multiple-choice questions. }
  \label{fig:choice_generation_prompt}
\end{figure}

\begin{figure}[t]
  \begin{tcolorbox}[colback=white, width=\textwidth, title=Result Extraction Instructions]
    \begin{lstlisting}[basicstyle=\tiny\ttfamily, breaklines=true]
    You are a biomedical research assistant with extensive expertise in conducting and reviewing observational studies. Your task is to systematically extract quantitative results from a study's abstract into a structured format. The goal is to establish ground-truth data that can be used to automatically evaluate the quality of third-party reproductions of the study.
    For each quantitative result, extract the information using the following fields:
    - `result_type` (str): The type of result. Must be one of:
        - `baseline_characteristics`: baseline characteristics of the study population or groups/cohorts. Attention: you should only extract the numerical values from experimental results, not hyperparameter settings from the experimental methods (e.g., database size, size of training/testing set, etc.)
        - `primary_results`: primary results of the study, including outcome statistics, effect sizes, and their confidence intervals or p-values.
        - `additional_results`: additional results such as subgroup analyses, sensitivity analyses, or other secondary findings.
    - `masked_description`(str): A one- to two-sentence description rewritten from the abstract. All numerical values of results must be replaced with short placeholder tags (e.g., <odds_ratio>, <risk_ratio>). The description must avoid anaphora (e.g., "A group of patients" -> specify what group).
    - `value_dict`(dict): A dictionary containing the actual extracted values. The keys must match the placeholder tags used in the masked_description, and the values should be the corresponding numbers (e.g., {"odds_ratio": "1.22", "lower_limit": "0.94", "upper_limit": "1.50"}).
    Each result should be assigned a unique index (integer), starting from 1, to indicate its order of appearance in the abstract.
    
    To ensure qualified and consistent output, you must strictly follow the task rules:
    1. ONLY the quantitative results should be extracted. If a masked_description contains no any numerical value result, it is NOT qualified.
    2. A single sentence may contain multiple results. If the results are directly related (e.g., an effect size and its confidence interval), extract them as one entry. If they are independent (e.g., different outcomes), rewrite the sentence and extract them separately as distinct entries.
    3. Extracted results must consist of numbers only, expressed in Arabic numerals (not words), and exclude units.
    4. The output must be a valid JSON object, where index is the key and the corresponding structured result is the value.
    
    An example is as follows:
    Abstract:
    [BACKGROUND] Fluid therapy is a cornerstone in the treatment of sepsis. Recently, the guidelines have recommended the combined administration that using crystalloids plus albumin for septic patients, but the optimal timing for albumin combined is still unclear. The objective of this study was to investigate the association of timing of albumin combined with 28-day mortality in patients with sepsis.\n[METHODS] We involved septic patients from the Medical Information Mart for Intensive Care (MIMIC)-IV database, and these patients were categorized into crystalloids group (crystalloids alone) and early combination group (crystalloids combined albumin at 0-24 h). The primary outcome was 28-day mortality. We used propensity score matching (PSM) to adjust confounding and restricted mean survival time (RMST) analysis was conducted to quantify the beneficial effect on survival due to the combination group.\n[RESULTS] We categorized 6597 and 920 patients in the \"crystalloids alone\" and \"early combination\", respectively. After PSM, compared to the crystalloids group, the combination group was associated with the increased survival among 28-day (increased survival: 3.39 days, 95% CI 2.53-4.25; P < 0.001) after ICU admission. Patients who received albumin combination at the first 24-h was associated with prolonged LOS in ICU (10.72 days vs. 8.24 days; P < 0.001) but lower risk of 28-day mortality (12.5% vs 16.4%, P = 0.003) than those received crystalloids alone.\n[CONCLUSION] In septic patients, receiving albumin combined within the first 24-h after crystalloids administration was associated with an increment of survival in 28 days.
    
    Output:
    ```json
    {
        "1": {
            "result_type": "baseline_characteristics",
            "masked_description": "We categorized <crystalloids_alone> and <early_combination> patients in the \"crystalloids alone\" and \"early combination\", respectively.",
            "value_dict": {"crystalloids_alone": "6597", "early_combination": "920"}
        },
        "2": {
            "result_type": "primary_results",
            "masked_description": "After PSM, compared to the crystalloids group, the combination group was associated with the increased survival among 28-day (<survival_days> days, 95% CI <lower_limit>-<upper_limit>; P < <p_value_boundary>) after ICU admission.",
            "value_dict": {"survival_days": "3.39", "lower_limit": "2.53", "upper_limit": "4.25", "p_value_boundary": "0.001"}
        },
        "3": {
            "result_type": "additional_results",
            "masked_description": "Patients who received albumin combination at the first 24-h was associated with prolonged LOS in ICU (<albumin_combination_days> days vs. <crystalloids_alone_days> days; P < <albumin_combination_p_value_boundary>) but lower risk of 28-day mortality (<mortality_rate_1>% vs <mortality_rate_2>%, P = <mortality_rate_p_value>) than those received crystalloids alone.",
            "value_dict": {"albumin_combination_days": "10.72", "crystalloids_alone_days": "8.24", "albumin_combination_p_value_boundary": "0.001", "mortality_rate_1": "12.5", "mortality_rate_2": "16.4", "mortality_rate_p_value": "0.003"}
        }
    }

    Please summarize the following study's results based on its abstract in a valid JSON format:
    Abstract:
    {abstract}
    \end{lstlisting}
  \end{tcolorbox}
    \cprotect
  \caption{Result extraction prompt. Variable \verb|{abstract}| will be replaced with publication's abstract at runtime.}
  \label{fig:result_extraction_prompt}
\end{figure}

\begin{figure}[t]
  \begin{tcolorbox}[colback=white, width=\textwidth, title=Prompt for Generating Field Types. Part 1 or 2]
    \begin{lstlisting}[basicstyle=\tiny\ttfamily, breaklines=true]
    You are an information-extraction and schema-annotation assistant for observational-study results.
    Your job: given an input JSON object whose keys are item ids (e.g., "1", "2", ...), where each item has:
    * masked_description (text with placeholders like <var>), and
    * value_dict (a dict of field -> numeric string),
    you must formulate a "field_schema" object under each item id.
    Each field_schema maps every key in value_dict to a schema entry used for downstream evaluation.
    
    Schema types and their facets contracts:
    In each "field_schema", include every key present in that item's value_dict. Do not add keys that are not in value_dict. Once you choose a 'type', you must supply that type's required facets (None if no given); include optional facets only if explicitly available or strongly implied (omit absent information).
    Choose exactly one 'type' per field from:
    * "count": integer-like counts (patients, cases, events).
        * Required facets: (none)
        * Optional facets: 'unit'
    * "proportion": probability / rate values (may be already normalized 0-1, or reported per-100/per-1000/per-100k; may be incidence per person-time).
        * Required facets: `proportion_base` belongs to {1, 100, 1000, 100000, None}
        * Optional facets: 
            * 'role' (e.g., "incidence_rate")
            * 'n_key' (sample size. binomial denominator field name)
            * 'denominator_key' (person-time field name)
            * 'unit' 
    * "ratio": effect ratios (HR, OR, RR)-multiplicative effect sizes.
        * Required facets: (none)
        * Optional facets:
            * 'ci_low_key' (confidence interval lower endpoint field name)
            * 'ci_high_key' (confidence interval upper endpoint field name)
            * 'role' (e.g., "odds_ratio", "risk_ratio", "hazard_ratio")
            * 'unit'
    * "numeric": general continuous values (e.g., means, medians, thresholds, cutoffs, dispersion).
        * Required facets: (none)
        * Optional facets:
            * `role` (e.g., "mean", "median", "threshold", "cutoff", "dispersion","sd","iqr")
            * `unit`
    * "auc": ROC AUC metrics (bounded [0,1]).
        * Required facets: (none)
        * Optional facets: (none)
    * "pvalue": p-values (exact, boundary, or interval endpoints)
        * Required facets: pvalue_mode belongs to {"exact", "bound", "interval_low", "interval_high"}
            * If pvalue_mode belongs to {"exact","bound"} -> require 'pvalue_sign' belongs to {"=", "<", "<=", ">", ">=", None} (choose based on the nearest sign on the left-hand side of the p-value placeholder in masked_description)
            * If pvalue_mode belongs to {"interval_low","interval_high"} -> require 'pair_key' (the other endpoint's field name) and omit 'pvalue_sign'.
        * Optional facets: (none)
    * "ci" : confidence interval endpoints (e.g., *_ci_lower, *_ci_upper).
        * Required facets: (none)
        * Optional facets: (none)
    * "other": catch-all when none of the above types fit
        * Required facets: (none)
        * Optional facets: `role`, `unit`

    An example is as follows:
    Input:
    {
      "1": {
        "masked_description": "A total of <total_patients> adult patients aged 18 years and older were enrolled in the study, with <men> men and <women> women included.",
        "value_dict": {
          "total_patients": "3273",
          "men": "1820",
          "women": "1453"
        }
      },
    \end{lstlisting}
  \end{tcolorbox}
    \cprotect
  \caption{Prompt for generating field types. Part 1 or 2.}
  \label{fig:field_type_prompt_1}
\end{figure}

\begin{figure}[t]
  \begin{tcolorbox}[colback=white, width=\textwidth, title=Prompt for Generating Field Types. Part 2 of 2]
    \begin{lstlisting}[basicstyle=\tiny\ttfamily, breaklines=true]
   
      "2": {
        "masked_description": "The incidence rates of in-hospital mortality and one-year mortality rate were <in_hospital_mortality_rate> per 1,000 person-days and <one_year_mortality_rate> per 1,000 person-years, respectively.",
        "value_dict": {
          "in_hospital_mortality_rate": "0.96",
          "one_year_mortality_rate": "263.8"
        }
      },
      "3": {
        "masked_description": "Multivariable regression analysis identified baseline FI_Lab > 0.45 as an independent risk factor predicting in-hospital mortality (odds ratio = <in_hospital_mortality_or>, 95% CI <in_hospital_mortality_ci_lower>-<in_hospital_mortality_ci_upper>, p < <in_hospital_mortality_p_value>) and one-year mortality (hazard ratio = <one_year_mortality_hr>, 95% CI: <one_year_mortality_ci_lower>-<one_year_mortality_ci_upper>, p < <one_year_mortality_p_value>).",
        "value_dict": {
          "in_hospital_mortality_or": "3.221",
          "in_hospital_mortality_ci_lower": "2.341",
          "in_hospital_mortality_ci_upper": "4.432",
          "in_hospital_mortality_p_value": "0.001",
          "one_year_mortality_hr": "2.152",
          "one_year_mortality_ci_lower": "1.730",
          "one_year_mortality_ci_upper": "2.678",
          "one_year_mortality_p_value": "0.001"
        }
      }
    }
    
    Output:
    ```json
    {
      "1": {
        "field_schema": {
          "total_patients": { "type": "count" },
          "men": { "type": "count" },
          "women": { "type": "count" }
        }
      },
      "2": {
        "field_schema": {
          "in_hospital_mortality_rate": {
            "type": "proportion",
            "role": "incidence_rate",
            "proportion_base": 1000
          },
          "one_year_mortality_rate": {
            "type": "proportion",
            "role": "incidence_rate",
            "proportion_base": 1000
          }
        }
      },
      "3": {
        "field_schema": {
          "in_hospital_mortality_or": {
            "type": "ratio",
            "ci_low_key": "in_hospital_mortality_ci_lower",
            "ci_high_key": "in_hospital_mortality_ci_upper"
          },
          "in_hospital_mortality_ci_lower": { "type": "ci" },
          "in_hospital_mortality_ci_upper": { "type": "ci" },
          "in_hospital_mortality_p_value": {
            "type": "pvalue",
            "pvalue_sign": "<"
          },
          "one_year_mortality_hr": {
            "type": "ratio",
            "ci_low_key": "one_year_mortality_ci_lower",
            "ci_high_key": "one_year_mortality_ci_upper"
          },
          "one_year_mortality_ci_lower": { "type": "ci" },
          "one_year_mortality_ci_upper": { "type": "ci" },
          "one_year_mortality_p_value": {
            "type": "pvalue",
            "pvalue_sign": "<"
          }
        }
      }
    }
    ```
    \end{lstlisting}
  \end{tcolorbox}
    \cprotect
  \caption{Prompt for generating field types. Part 2 of 2.}
  \label{fig:field_type_prompt_2}
\end{figure}

\begin{figure}[t]
  \begin{tcolorbox}[colback=white, width=\textwidth, title=System Prompt of MLAB and RWEAgent]
    \begin{lstlisting}[basicstyle=\tiny\ttfamily, breaklines=true]
    You are a highly capable medical data scientist specializing in real-world evidence (RWE) and observational study execution. You are proficient in R programming and have extensive experience applying R to medical data analysis.

    You have access to following actions with params, and receive corresponding feedback after each action:
    list_files: Lists files and directories in a specified path.
    * parameters: 
        - path (optional): str, a relative path to list contents from. Defaults to agent workspace.
    * return: A list of files and directories in the specified path.
    * valid response format to execute this action:
    {tools_desc}
    You MUST always follow the following rules:
    * Always ensure you fully understand the table structure of database you operate on and consider this structure before interacting with the database.
    * If you create multiple R scripts, use the source() function to connect their execution. For intermediate results that are large or time-consuming to compute, use saveRDS() and readRDS() to persist and reload them.
    * It is recommended to add print() or cat() statements at the end of each R script to inspect or illustrate sample data or summary statistics of the execution results if applicable.
    * Every your response MUST have ONLY ONE action call in valid JSON format and then wait for the environment feedback. Never fabricate or alter observations!
    **ALWAYS RESPOND IN THIS FORMAT EXACTLY**:
    Reflection: 
    (What does the observation mean? If there is an error, what caused the error and how to debug?)
    Research Plan and Status: 
    (The full high level research plan, with current status and confirmed results of each step briefly annotated. It must only include progress that has been made by previous steps. If there is any update, enclose the new update text in double asterisks **like this**. If there is no update, just copy the previous step Research Plan and Status. The high level plan from the previous step should be fully retained, unless it is intentionally revised.)
    Fact Check: 
    (List all objective statements in the updates to Research Plan and Status one by one and point out whether it is guessed versus directly confirmed by the previous observation directly above. Performance numbers can only be confirmed by running the code and observing the output)
    Thought: 
    (What you are currently doing, what actions to perform and why)
    Action: 
    (the JSON format action to take, should be ONLY ONE action selected from the available actions. NO AND MORE content after the action.)
    
    After taking an action, you will receive an observation of the environment.
    \end{lstlisting}
  \end{tcolorbox}
    \cprotect
  \caption{System prompt of MLAB and RWEAgent. Variable \verb|{tools_desc}| is replaced with corresponding toolset description based on the selected scaffold.}
  \label{fig:sys_prompt_MLAB_EHRAgent}
\end{figure}

\begin{figure}[ht]
  \begin{tcolorbox}[colback=white, width=\textwidth, title=System Prompt of Memory LLM]
    \begin{lstlisting}[basicstyle=\tiny\ttfamily, breaklines=true]
    You are a Data Analysis Research Assistant. Your task is to synthesize the operational history of a medical data scientist conducting an observational study in an R environment. The history consists of the scientist's thoughts, actions, and observations (environment feedback).
    
    The goal is to create a concise, structured "state summary" in JSON format. This summary will serve as the scientist's memory, allowing it to continue the analysis without needing the full verbose history. Therefore, the summary should reach the balance between being concise and operational, removing trivial details and emphasizing the key experiences and findings.
    
    JSON Output:
    {
      "progress_summary": "A summary of the overall progress made during the summarized period within five sentences.",
      "key_decisions": [
        "A list of critical choices made by the agent that altered the analysis path, e.g., 'Decided to use mean imputation for missing 'age' values instead of row deletion.'"
      ],
      "key_insights_and_findings": [
        "Crucial discoveries made *about the data*, e.g., 'Found a significant positive correlation (p < 0.05) between BMI and blood pressure.'"
      ],
      "technical_constraints_and_learnings": [
        "Important discoveries *about the environment, data schema, or tools*, e.g., 'The database table 'PatientRecords' lacks a unique 'row_id' column.', 'The `run_model()` function does not have a 'verbose' parameter.'"
      ],
      "artifact_status": [ # you should record every file's function, usage, and status if the scientist have interacted with it, so that the scientist can continue the analysis from the last state
        {
          "artifact_name": "e.g., 'DatabaseConfig.R'",
          "function": "A brief description of its function, e.g., 'This file is used to load the database connection details.'",
          "usage": "A brief description of its usage, e.g., 'Every R script should source this file to load the database connection details if it needs to connect to the database.'",
          "status": "A brief description of its current state, e.g., 'Completed and executed successfully.'"
        },
        {
          "artifact_name": "e.g., 'xgboost_v1.model'",
          "function": "A brief description of its function, e.g., 'This file is the XGBoost model parameters trained in xxx file.'",
          "usage": "A brief description of its usage, e.g., 'Use yyy file to load the model parameters to predict the outcome.'",
          "status": "e.g., 'Initial training shows poor performance (AUC < 0.6). The training script needs to be revised.'"
        }
      ],
      "open_questions_or_next_steps": [
        "The user still needs to provide information about Z.",
        "The agent's next action should be to investigate W."
      ]
    }
    \end{lstlisting}
  \end{tcolorbox}
    \cprotect
  \caption{System prompt of memory LLM.}
  \label{fig:memory_llm_sys_prompt}
\end{figure}

\begin{figure}[ht]
  \begin{tcolorbox}[colback=white, width=\textwidth, title=System Prompt of Cohort Judge]
    \begin{lstlisting}[basicstyle=\tiny\ttfamily, breaklines=true]
    Act as a medical doctor reviewing a patient's healthcare data captured during routine clinical care, such as electronic
    health records andinsurance claims.
    Your task is to determine whether a patient should be excluded from a study cohort, based strictly on the provided cohort definition.
    You will be given:
    * A cohort definition
    * Patient-level clinical information
    
    Instructions
    * Check each inclusion/exclusion criterion one by one against the patient data.
    * Use only explicit evidence from the provided information to decide whether the patient should be excluded.
    * Do not assume the presence or absence of conditions when information is missing. If there are some criteria that cannot be evaluated, default to being met and add it to the unevaluable_criteria list.
    * If any inclusion criterion is not satisfied, the patient should be excluded.
    * If any exclusion criterion is satisfied, the patient should be excluded.
    
    Output (JSON only)
    Return a single JSON object with the following fields:
    {
      "unevaluable_criteria": [],
      "decision": "Exclude | Do not exclude",
      "triggered_exclusion_criteria": [],
      "evidence": ""
    }
    * unevaluable_criteria: a list of inclusion or exclusion criteria (original wording) that cannot be evaluated given the provided patient information. Use an empty list if all criteria can be evaluated.
    * triggered_exclusion_criteria: a list of exclusion criteria (original wording) that apply; use an empty list if none apply.
    * evidence: a brief explanation citing relevant patient information. Only return if the patient is excluded else return "N/A".
    Do not include any text outside the JSON object. Do not modify or reinterpret the cohort definition.
    \end{lstlisting}
  \end{tcolorbox}
    \cprotect
  \caption{System prompt of Cohort Judge.}
  \label{fig:cohort_judge_prompt}
\end{figure}


\end{document}